\documentclass[pdflatex,sn-mathphys]{sn-jnl}

\theoremstyle{thmstyleone}%
\theoremstyle{thmstyletwo}%

\theoremstyle{thmstylethree}%

\usepackage{arydshln}
\usepackage[usestackEOL]{stackengine}
\usepackage{commath}
\newcommand{\specialcell}[2][c]{%
  \begin{tabular}[#1]{@{}c@{}}#2\end{tabular}}

\begin{document}

\title{Computer Vision Self-supervised Learning Methods on Time Series}


\author*[1]{\fnm{Daesoo} \sur{Lee}}
\author[1,2]{\fnm{Erlend} \sur{Aune}}

\affil*[1]{\orgdiv{Department of Mathematical Sciences}, \orgname{Norwegian University of Science and Technology}, \orgaddress{\city{Trondheim}, \country{Norway}}}

\affil[2]{\orgname{Abelee}, \orgaddress{\city{Oslo}, \country{Norway}}}




\abstract{Self-supervised learning (SSL) has had great success in both computer vision. Most of the current mainstream computer vision SSL frameworks are based on Siamese network architecture. These approaches often rely on cleverly crafted loss functions and training setups to avoid feature collapse. In this study, we evaluate if those computer-vision SSL frameworks are also effective on a different modality (\textit{i.e.,} time series). The effectiveness is experimented and evaluated on the UCR and UEA archives, and we show that the computer vision SSL frameworks can be effective even for time series. In addition, we propose a new method that improves on the recently proposed VICReg method. Our method improves on a \textit{covariance} term proposed in VICReg, and in addition we augment the head of the architecture by an iterative normalization layer that accelerates the convergence of the model. 
}

\keywords{Self-supervised learning, Time series, VICReg, VIbCReg}



\maketitle

\section{Introduction}\label{sec1}

In recent years, representation learning has had great success within computer vision, improving both on SOTA for fine-tuned models and achieving close-to SOTA results on linear evaluation on the learned representations \cite{Henaff2019Data-efficientCoding, chen2020simple, he2020momentum, grill2020bootstrap,Chen2020ExploringLearning,Zbontar2021BarlowReduction,Bardes2021VICReg:Learning}, and many more. The main idea in these papers is to train a high-capacity neural network using a self-supervised learning (SSL) loss with a Siamese network's architectural style \cite{Koch2015SiameseRecognition}. The SSL loss is able to produce representations of images that are useful for downstream tasks such as image classification and segmentation. 


Most mainstream SSL frameworks have been developed in computer vision, and SSL for time series problems has seen much less attention despite their evident abundance in industrial, financial, and societal applications. The Makridakis competitions \cite{makridakis2018m4, makridakis2020m5} give examples of important econometric forecasting challenges. The UCR archive \cite{UCRArchive} is a collection of univariate time series datasets where classification is of importance, and in the industrial setting, sensor data and Internet of Things (IoT) data are examples where proper machine learning tools for time series modeling is important. The UEA archive \cite{bagnall2018uea} was designed as a first attempt to provide a standard archive for multivariate time series classification similar to UCR. 

Our main focus is to test if the computer-vision SSL methods are also effective on time series so that we can potentially bring some of the quality ideas from the computer-vision representation learning literature to that of the time series. 
Thus, we investigate the mainstream SSL frameworks from computer vision and their effectiveness on time series datasets along with SSL frameworks for time series, and show that the computer-vision SSL frameworks are effective not only on images but also on time series. Also, we propose a new SSL framework named VIbCReg (Variance-Invariance-better-Covariance Regularization) which is based on VICReg and inspired by the feature decorrelation methods from \cite{Ermolov2020WhiteningLearning,Hua2021OnLearning,Bardes2021VICReg:Learning}. VIbCReg can be viewed as an upgraded version of VICReg by having better covariance regularization.

The recent mainstream SSL frameworks can be divided into two main categories: 1) contrastive learning method, 2) non-contrastive learning method. The representative contrastive learning methods such as MoCo \cite{he2020momentum} and SimCLR \cite{Chen2020ARepresentations} use positive and negative pairs and they learn representations by pulling the representations of the positive pairs together and pushing those of the negative pairs apart. However, these methods require a large number of negative pairs per positive pair to learn representations effectively. To eliminate the need for negative pairs, non-contrastive learning methods such as BYOL \cite{grill2020bootstrap}, SimSiam \cite{Chen2020ExploringLearning}, Barlow Twins \cite{Zbontar2021BarlowReduction}, and VICReg \cite{Bardes2021VICReg:Learning} have been proposed. Since non-contrastive learning methods use positive pairs only, their training setup could be simplified. Non-contrastive learning methods are also able to outperform the existing contrastive learning methods. 

To further improve the quality of the learned representations, feature whitening and feature decorrelation have been main ideas behind some recent improvements \cite{Ermolov2020WhiteningLearning,Zbontar2021BarlowReduction,Hua2021OnLearning,Bardes2021VICReg:Learning}. Initial SSL frameworks such as SimCLR suffer from a problem called feature collapse. When not enough negative pairs are available, the representations collapse to constant vectors. The collapse occurs since a similarity metric is still high even if all the features converged to constants, which is the reason for the use of the negative pairs to prevent feature collapse. Feature collapse has been partially resolved by using a momentum encoder \cite{grill2020bootstrap}, an asymmetric framework with a predictor, and stop-gradient \cite{grill2020bootstrap,Chen2020ExploringLearning}, which have popularized the non-contrastive learning methods. However, one of the latest SSL frameworks, VICReg, shows that none of them is needed and that it is possible to conduct effective representation learning with the simplest Siamese architecture without the collapse via feature decorrelation. Using this idea for SSL has first shown up in W-MSE \cite{Ermolov2020WhiteningLearning} recently, where the feature components are decorrelated from each other by whitening. Later, Barlow Twins encodes feature decorrelation by reducing off-diagonal terms of its cross-correlation matrix to zero. Hua et al. \cite{Hua2021OnLearning} encodes the feature decorrelation by introducing Decorrelated Batch Normalization (DBN) \cite{Huangi2018DecorrelatedNormalization} and Shuffled-DBN. VICReg handles feature decorrelation by introducing variance and covariance regularization terms in its loss function in addition to a similarity loss.

The computer-vision SSL frameworks investigated in this paper are SimCLR, BYOL, SimSiam, Barlow Twins, and VICReg along with our proposal, VIbCReg. 

The SSL frameworks are evaluated 1) on a subset of UCR by linear and fine-tuning evaluation, and 2) on the UCR and UEA archives with the SVM (support vector machine) evaluation suggested in \cite{franceschi2019unsupervised}. The SVM evaluation is conducted by fitting an SVM classifier on learned representations. As for the subset of UCR, ten datasets of the UCR archive with the highest number of time series within them and five datasets of the UCR archive with the highest number of labels are used.



\section{Related Works}
\begin{figure}[ht]
\centering
\includegraphics[width=\textwidth]{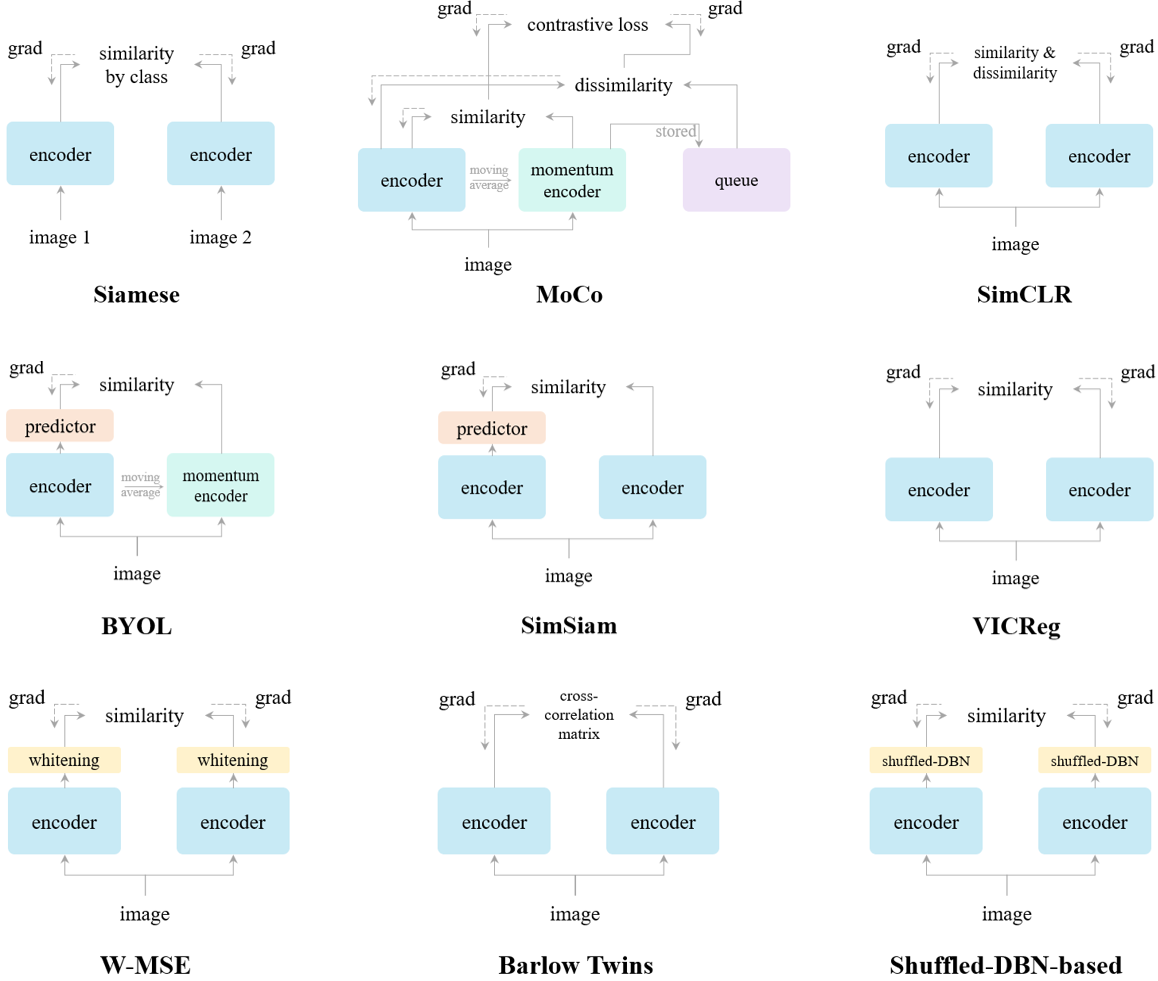}
\caption{Comparison on Siamese architecture-based SSL frameworks. Siamese denotes the Siamese architecture \cite{Koch2015SiameseRecognition} and the others are SSL frameworks. The encoder includes all layers that can be shared between both branches. The dash lines indicate the gradient propagation flow. Therefore, the lack of a dash line denotes stop-gradient.}
\label{fig:Siamese_based_SSL}
\end{figure}

\subsection{Contrastive Learning Methods}
\paragraph{Siamese Architecture-based SSL Frameworks}  
We illustrate in Fig. \ref{fig:Siamese_based_SSL} a comparison between several Siamese neural network \cite{Chopra2005LearningVerification} SSL frameworks.
The Siamese neural network has two branches with a shared encoder on each side, and its similarity metric is computed based on two representations from the two encoders. Its main purpose is to learn representations such that two representations are similar if two input images belong to the same class and two representations are dissimilar if two input images belong to different classes. One of the main applications of the Siamese neural network is one-shot and few-shot learning for image verification \cite{Chopra2005LearningVerification,Koch2015SiameseRecognition} where a representation of a given image can be obtained by an encoder of a trained Siamese neural network and is compared with representations of the existing images for the verification by a similarity metric. It should be noted that the Siamese neural network is trained in a supervised manner with a labeled dataset. The learning process of the representations by training an encoder is termed representation learning.

Although representation learning can be conducted using the Siamese neural network, its capability is somewhat limited by the fact that it utilizes supervised learning where a labeled dataset is required. To eliminate the need for a labeled dataset and still be able to conduct effective representation learning, SSL has become very popular where unlabeled datasets are utilized to train an encoder. Some of the representative contrastive learning methods that are based on the Siamese architecture are MoCo and SimCLR. Both frameworks are illustrated in Fig. \ref{fig:Siamese_based_SSL}.

\textbf{MoCo} is a contrastive learning method that requires a large number of negative pairs per positive pair, and MoCo keeps a large number of negative pairs by having a queue where a current mini-batch of representations is enqueued and the oldest mini-batch of representations in the queue is dequeued. A similarity loss is computed using representations from the encoder and the momentum encoder, and the dissimilarity loss is computed using representations from the encoder and the queue, and they are merged to form a form of a contrastive loss function, called InfoNCE \cite{VanDenOord2018RepresentationCoding}. The momentum encoder is a moving average of the encoder and it maintains consistency of the representations.

\textbf{SimCLR} greatly simplifies a framework for contrastive learning. Its major components are 1) a projector after a ResNet backbone encoder \cite{He2016DeepRecognition}, 2) InfoNCE based on a large mini-batch. The projector is a small neural network that maps representations to the space where contrastive loss is applied. By having this, quality of representations is improved by leaving the downstream task to the projector while the encoder is trained to output better quality representations.

\paragraph{SSL Frameworks for Time Series}


\textbf{TNC} is a recent SSL method to learn representations for non-stationary time series. It learns time series representations by ensuring that a distribution of signals from the same neighborhood is distinguishable from a distribution of non-neighboring signals. It was developed to address time series in the medical field, where modeling the dynamic nature of time series data is important.

\subsection{Non-Contrastive Learning Methods}
The representative non-contrastive learning methods are BYOL and SimSiam. Both frameworks are illustrated in Fig. \ref{fig:Siamese_based_SSL}.

\textbf{BYOL} has gained great popularity by proposing a framework that does not require any negative pair for the first time. Before BYOL, a large mini-batch size was required \cite{Chen2020ARepresentations} or some kind of memory bank \cite{Wu2018UnsupervisedDiscrimination,he2020momentum} was needed to keep a large number of negative pairs. Grill et al. \cite{grill2020bootstrap} hypothesized that BYOL may avoid the collapse without a negative pair due to a combination of 1) addition of a predictor to an encoder, forming an asymmetrical architecture and 2) use of the momentum encoder.

\textbf{SimSiam} can be viewed as a simplified version of BYOL where the momentum encoder is removed. Chen and He \cite{Chen2020ExploringLearning} empirically showed that a stop-gradient is critical to prevent the collapse.


\paragraph{Feature Decorrelation Considered}
The frameworks that improve the representation learning using the idea of feature decorrelation are W-MSE, Barlow Twins, and a Shuffled-DBN-based framework. They are illustrated in Fig. \ref{fig:Siamese_based_SSL}.

\textbf{W-MSE}'s core idea is to whiten feature components of representations so that the feature components are decorrelated from each other, which can eliminate feature redundancy. The whitening process used in W-MSE is based on a Cholesky decomposition \cite{Dereniowski2003CholeskyGraphs} proposed by Siarohin et al. \cite{Siarohin2019WhiteningGANS}. 

\textbf{Barlow Twins} encodes a similarity loss (invariance term) and a loss for feature decorrelation (redundancy reduction term) into the cross-correlation matrix. The cross-correlation is computed by conducting matrix multiplication of $z_i^T z_j$ where $z_i \in \mathbb{R}^{B \times F}$ and $z_j \in \mathbb{R}^{B \times F}$ and $B$ and $F$ denote batch size and feature size, respectively. Then, representation learning is conducted by optimizing the cross-correlation matrix to be an identity matrix, where optimization of the diagonal terms corresponds to that of a similarity loss and optimization of the off-diagonal terms corresponds to that of the feature decorrelation loss.

\textbf{DBN-based framework}'s authors \cite{Hua2021OnLearning} categorized the feature collapse into two categories: 1) \textit{complete collapse}, caused by constant feature components, 2) \textit{dimensional collapse}, caused by redundant feature components. They pointed out that previous works had mentioned and addressed the complete collapse only but not the dimensional collapse. Its main idea is to use DBN to normalize the output from its encoder-projector, which provides the feature decorrelation effect, and the feature decorrelation prevents the dimensional collapse. To further decorrelate the features, Shuffled-DBN is proposed, in which an order of feature components is randomly arranged before DBN and the output is rearranged back to the original feature-wise order. 

\paragraph{Feature Decorrelation and Feature-component Expressiveness Considered}
\hspace{\parindent}\textbf{VICReg} encodes \textit{Feature Decorrelation} (FD) and \textit{Feature-component Expressiveness} (FcE) in its loss function in addition to a similarity loss, where FD and FcE are termed \textit{variance term} and \textit{covariance term} in VICReg, respectively. A high FcE indicates that output values from a feature component have a high variance and vice versa. Hence, a (near)-zero FcE indicates the complete collapse. Its strength lies in its simplicity, VICReg uses the simplest Siamese form and does not require either the stop-gradient, asymmetrical architecture, or whitening/normalization layer. Despite its simplicity, its performance is very competitive compared to other latest SSL frameworks.

\subsection{Time Series Classification Methods}
In the time series classification (TSC) literature, \cite{malhotra2017timenet, franceschi2019unsupervised} propose scalable and transferable unsupervised representation learning frameworks and \cite{dempster2020rocket} proposes an unsupervised representation learning framework that is only scalable but not transferable. \cite{malhotra2017timenet} proposes TimeNet which is an autoencoder based on the seq2seq model \cite{sutskever2014sequence}. \cite{franceschi2019unsupervised} proposes a contrastive unsupervised learning method that uses a triplet loss. One of the main findings in TimeNet is that an encoder pretrained on many different datasets learns to effectively capture important characteristics of time series so that the pretrained encoder can also capture the important characteristics of time series of an unseen dataset. \cite{franceschi2019unsupervised} further improved a quality of learned representations by introducing a modified TCN (temporal convolutional network) \cite{lea2016temporal} to cover a large receptive field and a contrastive learning approach. \cite{franceschi2019unsupervised} also shows that its pretrained encoder has good transferability of its learned representations. Although ROCKET \cite{dempster2020rocket} produces representations that result in good classification performance, its representation size requires to be 20,000 which is extremely large compared to any of the typical representation learning methods (\textit{i.e.,} typically maximum 512) \cite{chen2020simple, he2020momentum, grill2020bootstrap,Chen2020ExploringLearning,Zbontar2021BarlowReduction,Bardes2021VICReg:Learning}. The large representation size limits a choice of a classifier. Also ROCKET's representations are not transferable from one dataset to another dataset unlike \cite{malhotra2017timenet, franceschi2019unsupervised}. Hence, TimeNet and the Franceschi are chosen to the main competing framework in our study.

\section{Proposed Method}
\label{section:proposed_method}
\begin{figure}[ht!]
\centering
\includegraphics[width=0.9\textwidth]{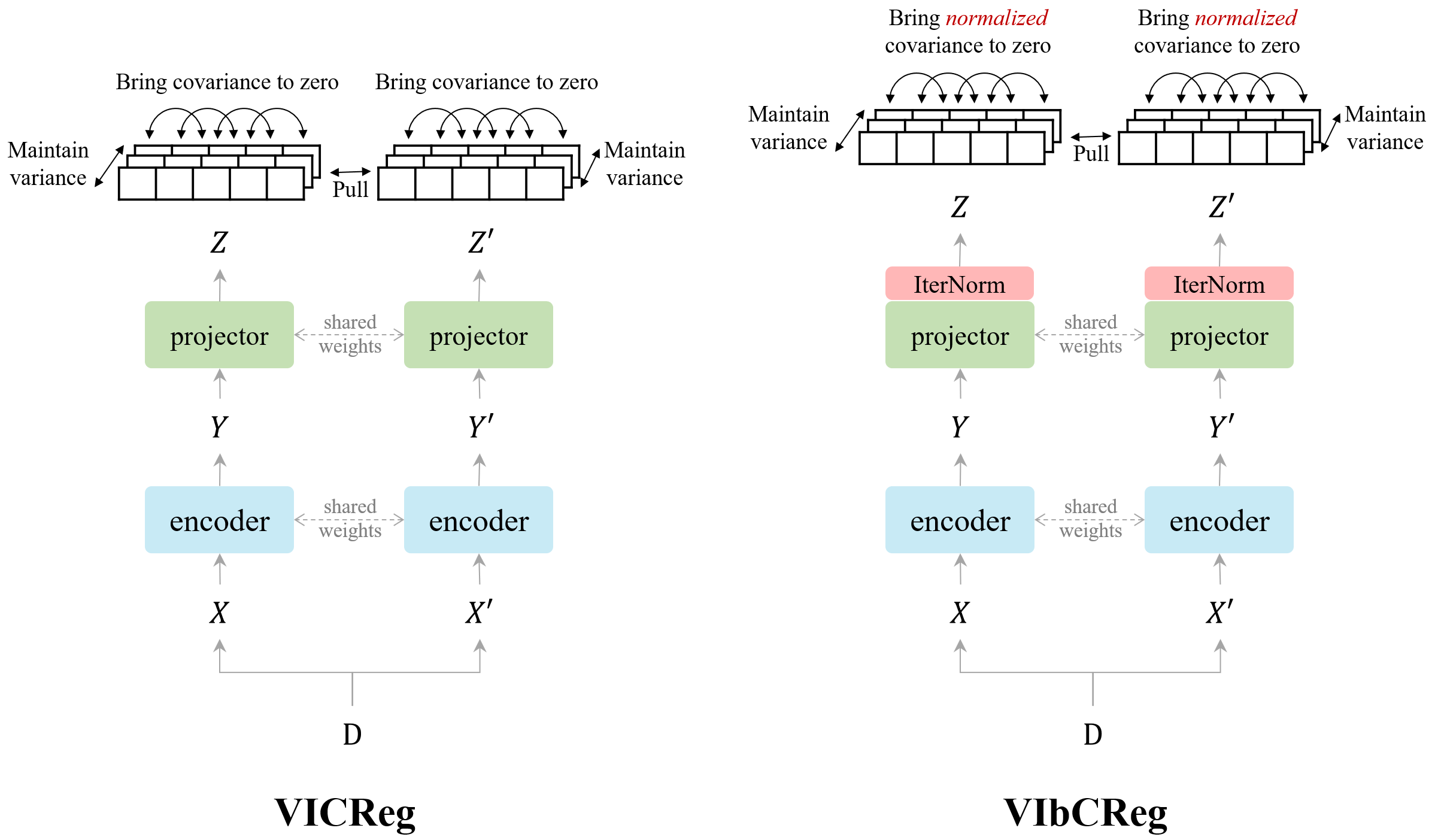}
\caption{Comparison between VICReg and VIbCReg, where the difference is highlighted in red. As for VIbCReg, two batches of different views $X$ and $X^\prime$ are taken from a batch of input data $\mathrm{D}$, and representations $Y$ and $Y^\prime$ are obtained through the encoder. Then, the representations are projected to a higher dimension via the projector and the iterative normalization (IterNorm) layer, yielding $Z$ and $Z^\prime$. Then, similarity between $Z$ and $Z^\prime$ and variance along the batch dimension are maximized, while feature components of $Z$ and $Z^\prime$ are decorrelated from each other.}

\label{fig:vibcreg}
\end{figure}
VIbCReg (Variance-Invariance-better-Covariance Regularization) is inspired by VICReg (Variance-Invariance-Covariance Regularization), and can be viewed as VICReg with a better covariance regularization. It should be noted that the variance, invariance, and covariance terms in VICReg correspond to what we call the \textit{feature component expressiveness} (FcE), similarity loss, and \textit{feature decorrelation} (FD) terms in VIbCReg, respectively. By having better covariance regularization, VIbCReg outperforms VICReg in terms of learning speed, linear evaluation, and semi-supervised training. What motivated VIbCReg are the followings: 1) FD has been one of the key components for improvement of quality of learned representations \cite{Ermolov2020WhiteningLearning,Zbontar2021BarlowReduction,Huangi2018DecorrelatedNormalization,Bardes2021VICReg:Learning}, 2) Hua et al. \cite{Hua2021OnLearning} showed that application of DBN and Shuffled-DBN on the output from an encoder-projector is very effective for faster and further FD, 3) addressing FcE in addition to FD is shown to be effective \cite{Bardes2021VICReg:Learning}, 4) scale of the VICReg's FD loss (covariance term) varies depending on feature size and its scale range is quite wide due to its summation over the covariance matrix. Hence, we assumed that the scale of the VICReg's FD loss could be modified to be consistent and have a small range so that a weight parameter for the FD loss would not have to be re-tuned in accordance with a change of the feature size.

VIbCReg is illustrated in Fig. \ref{fig:vibcreg} in comparison to VICReg. In Fig. \ref{fig:vibcreg}, 
the two views, $X$ and $X^{\prime}$, are encoded using the encoder into representations $Y$ and $Y^{\prime}$. The representations are further processed by the projector and an iterative normalization (IterNorm) layer \cite{Huang2019IterativeWhitening} into projections $Z$ and $Z^{\prime}$. The loss is computed at the projection level on $Z$ and $Z^{\prime}$.

We describe here a loss function of VIbCReg, which consists of a similarity loss, FcE loss, and FD loss. It should be noted that the similarity loss and the FcE loss are defined the same as in VICReg. The input data is processed in batches, and we denote $Z = [z_1, ..., z_B]^T \in \mathbb{R}^{B \times F}$ and $Z^{\prime} = [z_1^{\prime}, ..., z_B^{\prime}]^T \in \mathbb{R}^{B \times F}$ where $B$ and $F$ denote batch size and feature size, respectively. The \textit{similarity loss} (\textit{invariance term}) is defined as Eq. (\ref{eq:similarity_loss}). The \textit{FcE loss} (\textit{variance term}) is defined as Eq. (\ref{eq:variance_loss}), where $\mathrm{Var(..)}$ denotes a variance estimator, $\gamma$ is a target value for the standard deviation, fixed to 1 in our experiments, $\epsilon$ is a small scalar (\textit{i.e.,} 0.0001) preventing numerical instabilities.

\begin{equation}
    s(Z, Z^{\prime}) = \frac{1}{B} \sum_{b=1}^B \| Z_b - Z_b^{\prime} \|_2^2
    \label{eq:similarity_loss}
\end{equation}

\begin{equation}
    v(Z) = \frac{1}{F} \sum_{f=1}^F \mathrm{ReLU}\left( \gamma - \sqrt{\mathrm{Var}\left( Z_f \right) + \epsilon} \right)
    \label{eq:variance_loss}
\end{equation}

The \textit{FD loss} is the only component of VIbCReg's loss that differs from the corresponding \textit{covariance term} in the loss of VICReg, thus, we present here a comparison between the two. The covariance matrix of $Z$ in VICReg is defined as Eq. (\ref{eq:cov_mat_vicreg}), and the covariance matrix in VIbCReg is defined as Eq. (\ref{eq:cov_mat_vibcreg}) where the $\ell_2$-norm is conducted along the batch dimension. Then, the FD loss of VICReg is defined as Eq. (\ref{eq:FD_loss_vicreg}) and that of VIbCReg is defined as Eq. (\ref{eq:FD_loss_vibcreg}). Eq. (\ref{eq:cov_mat_vibcreg}) constrains the terms to range from -1 to 1, and Eq. (\ref{eq:FD_loss_vibcreg}) takes a mean of the covariance matrix by dividing the summation by a number of the matrix elements. Hence, Eq. (\ref{eq:cov_mat_vibcreg}) and Eq. (\ref{eq:FD_loss_vibcreg}) keep a scale and a range of the FD loss neat and small.

\begin{equation}
    C(Z)_{\mathrm{VICReg}} = \frac{1}{B-1} \left(Z - \bar{Z}\right)^T \left(Z - \bar{Z} \right) \: \: \mathrm{where} \; \bar{Z}=\frac{1}{B}\sum_{b=1}^{B}{Z_b}
    \label{eq:cov_mat_vicreg}
\end{equation}

\begin{equation}
    C(Z)_{\mathrm{VIbCReg}} = \left( \frac{Z - \bar{Z}}{\| Z - \bar{Z} \|_2} \right)^T \left( \frac{Z - \bar{Z}}{\| Z - \bar{Z} \|_2} \right)   
    \label{eq:cov_mat_vibcreg}
\end{equation}


\begin{equation}
    c(Z)_{\mathrm{VICReg}} = \frac{1}{F} \sum_{i \neq j} C_{\mathrm{VICReg}}(Z)^2_{i, j}
    \label{eq:FD_loss_vicreg}
\end{equation}

\begin{equation}
    c(Z)_{\mathrm{VIbCReg}} = \frac{1}{F^2} \sum_{i \neq j} C_{\mathrm{VIbCReg}}(Z)^2_{i, j}
    \label{eq:FD_loss_vibcreg}
\end{equation}

The overall loss function is a weighted average of the similarity loss, FcE loss, and FD loss:
\begin{equation}
    l (Z, Z^{\prime}) = \lambda s(Z, Z^{\prime}) + \mu\{ v(Z) + v(Z^{\prime}) \} + \nu\{ c(Z) + c(Z^{\prime}) \}
    \label{eq:overall_loss}
\end{equation}
where $\lambda$, $\mu$, and $\nu$ are hyper-parameters controlling the importance of each term in the loss.

Another key component in VIbCReg is \textit{IterNorm}. \cite{Hua2021OnLearning} showed that applying DBN on the output from an encoder-projector could improve representation learning, emphasizing that the whitening process helps the learning process, which corresponds to a main argument in the DBN paper \cite{Huangi2018DecorrelatedNormalization}. To further improve DBN, IterNorm was proposed \cite{Huang2019IterativeWhitening}. IterNorm was verified to be more efficient for whitening than DBN by employing Newton's iteration to approximate a whitening matrix. Therefore, we employ IterNorm instead of DBN, and IterNorm is applied to the output from an encoder-projector as shown in Fig. \ref{fig:vibcreg}. Then, feature decorrelation is strongly supported, induced by the following two factors: 1) IterNorm, 2) optimization of the FD loss. 

\paragraph{Relation to VICReg} To summarize, VIbCReg can be viewed as VICReg with the normalized covariance matrix (instead of the covariance matrix) and IterNorm. 

\begin{figure}[ht!]
\centering
\includegraphics[width=1.\textwidth]{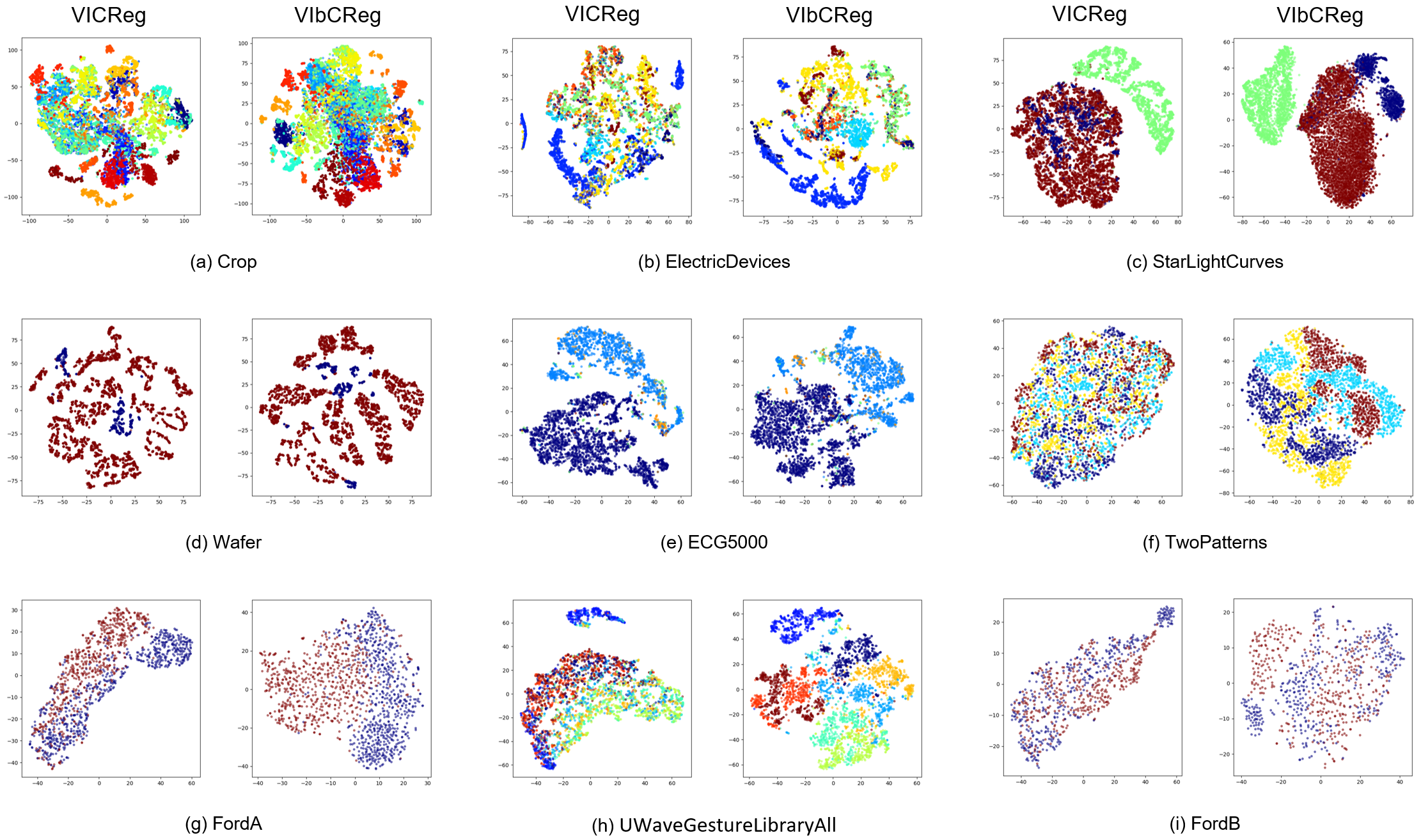}
\caption{T-SNE visualization of learned representations with VICReg and VIbCReg, respectively. For each set of subfigures, the left subfigure is by VICReg and the right subfigure is by VIbCReg. A subset dataset name is shown below each set of the subfigures and the different colors represent different classes.}
\label{fig:tsne_vicreg_vibcreg}
\end{figure}

Theoretically, VIbCReg should produce representations that are better decorrelated. To show that, we present T-SNE visualization of learned representations with VICReg and VIbCReg for several datasets from the UCR archive in Fig. \ref{fig:tsne_vicreg_vibcreg}. For (a)-(f) and (h), it is apparent that VIbCReg's representations are better in terms of class-separability. For (g) and (i), VIbCReg's representations are more scattered out due to the stronger decorrelation. It may seem that VICReg's representations are better separated but when looked closely, VICReg's representations are quite mixed between two classes around the boundary line between the two classes unlike VIbCReg's. This T-SNE result is well aligned with the linear evaluation result presented in Table \ref{table:LE_UCR}.

\section{Experimental Evaluation}
\label{section:experimental_evaluation}

Our experimental evaluation consists of two parts: 1) evaluation according to the common protocol for SSL frameworks in computer vision, 2) The SVM evaluation \cite{franceschi2019unsupervised}. In the part-1 evaluation, SimCLR, BYOL, SimSiam, Barlow Twins, VICReg, TNC, and VIbCReg are evaluated on a subset of the UCR archive. In the part-2 evaluation, all the computer-vision SSL frameworks (\textit{e.g.,} SimCLR, BYOL, SimSiam, Barlow Twins, VICReg), VIbCReg are evaluated on the UCR and UEA archives compared to the Franceschi.

\paragraph{Encoder} We follow the convention of the previous SSL papers by using ResNet \cite{grill2020bootstrap, Chen2020ExploringLearning, Zbontar2021BarlowReduction, Bardes2021VICReg:Learning} for an encoder. But since the dataset size is much smaller than the size for SSL in computer vision, we use a \textit{light-weighted} 1D ResNet, inspired by \cite{github_multi_scale_1d_resnet}. The detailed architecture is illustrated in Appendix \ref{fig:resnet}. It is used as an encoder for SimCLR, BYOL, SimSiam, Barlow Twins, VICReg, TNC, and VIbCReg.

\paragraph{Training} We use the AdamW \cite{LoshchilovDECOUPLEDREGULARIZATION} (\textit{lr}=0.001, weight decay=0.00001, batch size=256) optimizer, and 100 epochs of pretraining for the part-1 evaluation and 200 epochs for the part-2 evaluation. We use a cosine learning rate scheduler \cite{Loshchilov2017SGDR:Restarts}. Details of the data augmentation methods are described in Appendix \ref{appendix:data_augmentation_methods}.
We use PyTorch \cite{NEURIPS2019_9015} and train our models on a single GPU (GTX1080-Ti).

\subsection{Experimental Evaluation: Part 1}

\paragraph{Datasets} In this evaluation, the UCR archive is used, from which we select the 10 largest datasets together with the 5 datasets with more than 900 samples having the largest number of classes. For the UCR datasets, each dataset is split into 80\% (training set) and 20\% (test set) by a stratified split. A model pre-training by SSL is conducted and evaluated on each dataset. Therefore, the UCR datasets are independent of one another in our experiments. 
As for the preprocessing, the UCR datasets are preprocessed by z-normalization followed by arcsinh.
Summary of the UCR datasets is presented in Table \ref{tab:summary_ucr}.

\begin{linespread}{1.0}
\begin{table}[ht]
    \centering
    \caption{Summary of the UCR datasets. The first 10 datasets are the 10 largest datasets from the UCR archive, and the second 5 datasets are the datasets with the 5 largest number of classes from the UCR archive, where a dataset has more than 900 samples.}
    \label{tab:summary_ucr}
    \small
    \begin{tabular}{l c c c}
        \toprule 
        Dataset name & \#Samples & \#Classes & Length  \\
        \midrule
        Crop & 24000 & 24 & 46 \\
        ElectricDevices & 16637 & 7 & 96 \\
        StarLightCurves & 9236 & 3 & 1024 \\
        Wafer & 7164 & 2 & 152 \\
        ECG5000 & 5000 & 5 & 140 \\
        TwoPatterns & 5000 & 4 & 128 \\
        FordA & 4921 & 2 & 500 \\
        UWaveGestureLibraryAll & 4478 & 8 & 945 \\
        FordB & 4446 & 2 & 500 \\
        ChlorineConcentration & 4307 & 3 & 166 \\
        \hdashline
        ShapesAll & 1200 & 60 & 512 \\
        FiftyWords & 905 & 50 & 270 \\
        NonInvasiveFetalECGThorax1 & 3765 & 42 & 750 \\
        Phoneme & 2110 & 39 & 1024 \\
        WordSynonyms & 905 & 25 & 270 \\
        \botrule
    \end{tabular}
\end{table}
\end{linespread}


\paragraph{Linear Evaluation}
\label{subsection:LE}
We follow the linear evaluation protocols from the computer vision field \cite{grill2020bootstrap, Chen2020ExploringLearning, Zbontar2021BarlowReduction, Bardes2021VICReg:Learning, Koohpayegani2021MeanLearning}. Given the pre-trained encoder, we train a supervised linear classifier on the frozen features from the encoder. The features are from the ResNet's global average pooling (GAP) layer.
Experimental results for the linear evaluation on the UCR datasets are presented in Table \ref{table:LE_UCR}. 
For the linear classifier's training, the same optimizer and the same learning rate scheduler are used with training epochs of 50. Details of the data augmentation methods are described in Appendix \ref{appendix:data_augmentation_methods}.

The linear evaluation result on the UCR subset datasets shows that VIbCReg performs the best among the competing SSL frameworks and it even performs very close or outperforms the supervised one on many of the datasets, which indicates that quality of the learned representations by VIbCReg is very good as shown in Fig. \ref{fig:tsne_vicreg_vibcreg}. The second best is SimCLR, which is interesting given that SimCLR was proposed before BYOL, SimSiam, Barlow Twins, and VICReg. The main difference of SimCLR from other methods such as BYOL, SimSiam, Barlow Twins, and VICReg is that it is a contrastive method, while the others are contrastive methods. A contrastive method usually induces stronger decorrelation by explicitly pushing apart between one sample and the rest of the samples in a batch in the representation space. The effect can be observed in Fig. \ref{fig:UCR_FD_metrics}, where SimCLR generally achieves low FD metric. The FD metric is for measuring the feature decorrelation. The low FD metric indicates high feature decorrelation. In the same sense, VIbCReg also provides strong feature decorrelation. Thus, we suppose that strong feature decorrelation is important for linear class-separability of learned representations.

\begin{sidewaystable}
\sidewaystablefn%
\begin{center}
\begin{minipage}{\textheight}
\caption{Linear evaluation on the UCR datasets. The results are obtained over 5 runs with different random seeds for the stratified split. It is noticeable that the proposed frameworks outperform the other frameworks with significant margins, and they even outperform \texttt{Supervised} on some datasets. The values within the parentheses are standard deviations. Note that \texttt{Rand Init} denotes a randomly-initialized frozen encoder with a linear classifier on the top and \texttt{Supervised} denotes a trainable encoder-linear classifier trained in a supervised manner.}
\label{table:LE_UCR}
\tiny
\begin{tabular*}{\textheight}{@{\extracolsep{\fill}}lcccccccc  c  c@{\extracolsep{\fill}}}
\toprule%
Dataset Name & \texttt{Rand Init} & SimCLR & BYOL &  SimSiam  & Barlow Twins & VICReg & TNC  & \textbf{VIbCReg} & \texttt{Supervised} \\
\midrule
Crop & 49.6(0.1) & 65.6(1.2) & \underline{67.8(0.8)}  & 56.0(0.8)  & 63.7(0.7)  & 66.2(9.5) &  61.6(2.1)   & \textbf{71.0(0.7)}   & 80.1(0.4)  \\
ElectricDevices  & 51.2(0.7) & \textbf{87.7(0.4)} & 83.1(1.1) & 53.2(4.3)     & 64.1(1.1)     & 73.6(0.1) & 69.3(1.1) & \underline{87.1(0.3)}   & 87.0(0.2)     \\
StarLightCurves & 76.7(2.3) & \underline{97.4(0.4)} & 97.7(0.1) & 71.3(8.5)     & 88.7(4.2)     & 97.5(0.1) & 97.0(0.7)  & \textbf{97.8(0.1)}       & 98.3(0.1)     \\
Wafer  & 89.4(0.0)  & 93.2(1.2) &  \underline{99.4(0.5)} & 98.4(0.4) & 95.9(0.4)     & 98.8(0.2) & \textbf{99.5(0.0)} & \textbf{99.5(0.1)}       & 99.9(0.0)     \\
ECG5000     & 72.9(11.5) & \underline{94.8(0.1)} &  94.1(0.2) & 83.1(5.5) & 90.9(1.2)     & 92.8(0.0) & 93.3(0.3) & \textbf{95.4(0.1)}     & 95.8(0.1)     \\
TwoPatterns & 42.8(1.6)  & \textbf{99.4(0.2)} &  69.4(18.1) & 37.9(4.4) & 87.2(6.5)     & 81.2(0.6) & \underline{92.1(1.4)} & \textbf{99.3(0.2)}      & 100.0(0.0)    \\
FordA       & 54.5(0.9)  & \underline{95.1(0.0)} &  93.6(0.2) & 83.0(4.1)     & 74.5(4.3)     & 79.0(0.3) & 72.0(3.5) & \textbf{95.5(0.3)}       & 93.3(0.3)     \\
UWaveGestureLibraryAll & 47.2(1.3)  & 86.3(1.2) &  \underline{89.7(1.0)} & 30.3(6.8)     & 51.3(5.6)     & 57.5(0.8) & 64.8(1.6) & \textbf{90.9(0.4)} & 96.2(0.4)     \\
FordB                  & 65.7(1.1)  & 90.5(4.9) & \underline{94.0(0.2)} & 60.8(5.4) & 76.1(1.6) & 85.4(0.3) & 64.5(3.5) & \textbf{94.0(0.3)}      & 92.4(0.5)     \\
ChlorineConcentration  & 53.6(0.0)  & \underline{62.1(1.9)} & 57.4(0.4) & 55.7(0.0)  & 55.5(0.3) & 55.5(0.1) & 55.3(0.8) & \textbf{65.2(0.7)}     & 100.0(0.0)    \\
\hdashline
ShapesAll   & 7.9(2.5)  & \underline{80.7(3.4)} &  70.8(1.5) & 14.1(3.1)     & 39.2(3.6)   & 31.2(2.4)  & 51.8(4.5) & \textbf{85.7(0.8)}     & 91.2(1.0)     \\
FiftyWords  & 13.6(1.7)  & \underline{46.8(2.2)} &  30.0(0.3) & 15.8(2.1)  & 25.4(1.7)   & 26.3(0.9) & 26.2(2.6) & \textbf{50.2(1.1)}    & 77.2(1.2)     \\
NonInvasiveFetalECGThorax1 & 5.0(0.8)  & 51.9(3.5) &  \textbf{60.8(10.6)}  & 20.0(3.6)  & 21.4(8.8)   & 37.6(0.7) & 58.0(4.5) & \underline{58.5(0.6)}    & 94.5(0.3)     \\
Phoneme                    & 11.1(0.0)  & \underline{41.3(0.1)} &  38.9(1.1)  & 18.1(0.6)  & 19.9(1.5) & 21.4(0.2) & 28.0(2.2) & \textbf{42.8(0.5)} & 47.8(1.0)     \\
WordSynonyms               & 22.1(0.9)  & \textbf{43.8(0.5)} &  29.1(0.3) & 24.4(0.4)     & 29.0(1.3)   & 23.8(0.3) & 28.9(2.5) & \underline{46.7(0.7)}   & 73.7(1.8)     \\

\midrule
Mean Rank & 7.7 & 2.7 & 2.8 &	6.7 &	5.5	& 4.7	& 4.7	& 1.3  \\

\botrule

\end{tabular*}
\end{minipage}
\end{center}
\end{sidewaystable}

\paragraph{Fine-tuning Evaluation on a Small Dataset}
Fine-tuning evaluation on a subset of training dataset is conducted to do the ablation study. We follow the ablation study protocols from the computer vision domain \cite{grill2020bootstrap, Chen2020ExploringLearning, Zbontar2021BarlowReduction, Bardes2021VICReg:Learning, Koohpayegani2021MeanLearning}. Given the pre-trained encoder, we fine-tune the pre-trained encoder and train the linear classifier. Experimental results for the fine-tuning evaluation on the UCR datasets are presented in Table \ref{table:fine_tune_ev_UCR}.
AdamW optimizer is used with ($lr_{enc}$=0.0001, $lr_{cls}$=0.001, batch size=256, weight decay=0.001, training epochs=100), where $lr_{enc}$ and $lr_{cls}$ denote learning rates for the encoder and the linear classifier, respectively. During the fine-tuning, the BN statistics are set such that they can be updated. 
Details of the data augmentation methods are described in Appendix \ref{appendix:data_augmentation_methods}.

Similar to the linear evaluation result on the UCR, VIbCReg performs the best among the competing SSL frameworks in the fine-tuning evaluation on the UCR and beats the supervised one in the small-dataset regimes. 
Although SimCLR is the second best in the linear evaluation, it is not in the fine-tuning evaluation. VICReg performs the second best here. This indicates that VICReg's learned representations are easily adaptable to incoming data. The order mismatch in between linear evaluation and fine-tuning evaluation can also be found in \cite{xie2022simmim}, therefore, it is not uncommon to observe.

\begin{sidewaystable}
\sidewaystablefn%
\begin{center}
\begin{minipage}{\textheight}

\caption{Fine-tuning evaluation on subsets of the UCR datasets. The results are obtained over 5 runs with different random seed for the stratified split. In 5\% and 10\%, results on some datasets missing because the split training subsets are too small.}
\label{table:fine_tune_ev_UCR}
\tiny
\begin{tabular*}{\textheight}{@{\extracolsep{\fill}}lccccccc  c @{\extracolsep{\fill}}}
\toprule%
Dataset name & SimCLR & BYOL & SimSiam & Barlow Twins & VICReg & TNC & \textbf{VIbCReg}  & \texttt{Supervised} \\
\midrule
\multicolumn{7}{c}{\textit{Fine-tuning evaluation on 5\% of the training dataset}} \\
\midrule
Crop                   & 47.7(1.1) & 59.3(0.3) & 61.6(0.6)    & 60.5(0.5)   & 61.6(2.4) & 50.0(2.1)  & \underline{62.4(0.3)}        & \underline{62.6(0.7)}       \\
ElectricDevices        & \underline{82.1(0.7)} & 74.5(1.2) & 71.4(2.8)    & 69.2(1.5)   & 73.1(0.3) & 52.8(2.7) & \textbf{84.7(0.4)}     & 69.0(0.9)       \\
StarLightCurves        & 91.6(1.1) & 95.0(1.7) & 85.3(0.1)    & 93.0(4.3)   & \textbf{98.1(0.1)} & 85.9(0.5) & \textbf{98.1(0.1)}   & \underline{97.8(0.2)}       \\
Wafer                  & 89.3(0.1) & 90.3(1.5) & \underline{99.3(0.1)}    & 98.8(0.1)   & \textbf{99.4(0.1)} & 97.6(0.8) & {99.1(0.2)}   & \underline{99.3(0.2)} \\
ECG5000                & 92.8(0.2) & 91.9(1.0) & 91.4(1.2)    & 91.8(0.8)   & \textbf{94.0(0.5)} & 91.4(1.3) & \textbf{94.1(0.6)}  & \textbf{94.1(0.4)}       \\
TwoPatterns            & 91.6(0.5) & 37.1(8.4) & 49.3(8.5)    & 96.2(2.2)   & \underline{98.6(0.4)} & 66.0(3.7) & \textbf{99.2(0.5)}    & 88.0(4.2)       \\
FordA                  & \textbf{94.4(0.9)} & \underline{93.6(1.0)} & {91.1(1.0)}    & 80.6(3.5)   & 90.1(0.6) & 66.8(5.3)  & \textbf{94.4(0.6) }    & 89.8(1.0)       \\
UWaveGestureLibraryAll & 70.9(1.8) & 75.5(2.0) & 42.9(11.3)    & 55.1(4.6)   & 77.1(2.2) & 53.3(1.6) & \textbf{83.4(2.3)}     & \underline{78.5(1.1)}       \\
FordB                  & \underline{91.8(0.7)} & {91.2(0.4)} & 74.3(5.3)    & 82.4(2.1)   & {90.0(0.4)} & 60.5(5.3) & \textbf{92.2(0.7)}    & 87.4(1.9)       \\
ChlorineConcentration  & 52.7(1.0) & 52.0(2.4) & 57.4(0.7)    & 55.9(0.5)   & 56.7(1.0) & 54.7(0.6) & \underline{61.7(1.7)}    & \textbf{65.6(1.9)}       \\
\hdashline
NonInvasiveFetalECGThorax1 & 19.2(2.3) & 40.5(13.0) & 33.4(2.5)    & 21.2(6.1)   & 44.2(2.6) & 29.4(3.5)  & \underline{45.6(1.9)}    & \textbf{66.9(3.8)}       \\
\midrule
\multicolumn{7}{c}{\textit{Fine-tuning evaluation on 10\% of the training dataset}} \\
\midrule
Crop                     & 54.4(0.8) & 62.3(1.1) & \underline{66.2(0.6)}    & 65.3(0.4)         & \underline{66.4(1.9)} & 54.4(2.9)  & \underline{66.3(0.4)}    & \underline{67.0(0.7)}       \\
ElectricDevices          & \underline{83.3(0.3)} & 77.2(1.6)  & 75.1(1.9)    & 73.8(0.8)         & 75.9(0.5) & 60.9(2.5)  & \textbf{86.5(0.6)}  & 73.4(0.7)       \\
StarLightCurves          & 94.8(0.1) & 96.6(0.6)  & 91.8(5.3)    & 97.4(0.3)         & \textbf{98.2(0.1)} & 88.3(1.7)   & \textbf{98.2(0.1)}  & \underline{98.0(0.2)}       \\
Wafer                    & 89.3(0.2) & 93.1(3.3)  & \textbf{99.6(0.1)}    & \underline{99.4(0.1)}         & \textbf{99.6(0.1)} & 98.6(0.3)  & \textbf{99.5(0.1)}   & \textbf{99.6(0.1)}       \\
ECG5000                  & 93.3(0.1) & 92.4(0.1)  & 93.2(0.3)    & 93.1(0.5)         & \underline{94.6(0.7)} & 91.7(0.9)  & \textbf{95.0(0.5)}    & 94.4(0.7)       \\
TwoPatterns              & 94.9(0.7) & 41.4(10.2)  & 93.0(5.0)    & 99.5(0.5)         & \underline{99.7(0.1)} & 76.9(4.0) & \textbf{99.9(0.1)}    & \textbf{99.9(0.1)}       \\
FordA                    & \underline{94.5(0.4)} & {93.1(0.7)}  & 92.4(0.1)    & 87.5(1.4)         & {92.8(0.3)} & 68.8(3.5)  & \textbf{94.8(0.2)}   & 91.7(0.3)       \\
UWaveGestureLibraryAll   & 75.5(1.4) & 79.4(1.9)  & 58.1(14.0)    & 68.5(3.2)         & \underline{84.9(1.0)} & 55.7(3.7)  & \textbf{91.2(1.2)}   & \underline{84.7(0.9)}       \\
FordB                    & \underline{92.1(0.9)} & \underline{91.1(1.1)}  & 89.3(0.6)    & 89.3(1.9)  & {91.3(0.6)} & 63.3(4.9)  & \textbf{93.0(0.5)}  & 89.1(0.3)       \\
ChlorineConcentration    & 54.4(1.6) & 55.7(0.9)  & 63.4(0.7)    & 56.2(0.3)         & 60.6(1.3) & 55.4(0.2)  & \underline{72.2(2.6)}     & \textbf{76.7(2.2)}       \\
\hdashline
ShapesAll                  & \underline{57.5(4.5)} & 42.2(2.5) & 21.8(6.2)    & 35.4(3.4)         & 38.2(2.5) & 25.8(1.9)  & \textbf{67.8(3.1)}   & 50.2(4.0)       \\
NonInvasiveFetalECGThorax1 & 20.5(4.0) & 31.3(12.6) & 38.1(5.6)    & 24.4(3.7)         & 45.7(1.9) & 12.2(0.9)  & \underline{61.8(1.6)}   & \textbf{77.4(4.1)}       \\
WordSynonyms               & 37.9(4.5) & 28.5(4.2) & 29.3(1.1)    & 34.4(2.5)         & \underline{40.3(1.4)} & 27.6(6.0)  & \textbf{45.4(2.6)}    & 36.2(2.4)       \\
\midrule
\multicolumn{7}{c}{\textit{Fine-tuning evaluation on 20\% of the training dataset}} \\
\midrule
Crop                     & 58.2(0.7) & 64.6(0.7)  & \underline{70.4(0.5)}    & \underline{70.2(0.4)}    & \underline{70.4(1.3)} & 27.5(2.5)  & \underline{70.6(0.5)}     & \textbf{71.1(0.6)}  \\
ElectricDevices          & 58.7(0.0) & 78.7(1.3)  & \underline{80.4(0.9)}    & 79.7(0.5)    & 79.0(0.2) & 64.5(2.8)  & \textbf{88.1(0.3)}      & 78.5(0.6)       \\
StarLightCurves          & 96.7(0.2) & 97.6(0.3)  & 98.0(0.3)    & 98.0(0.2)    & \textbf{98.4(0.1)} & 93.1(2.5)  & \textbf{98.3(0.1)}     & \underline{98.2(0.1)}       \\
Wafer                    & 89.4(0.3) & 97.8(0.6)  & \textbf{99.7(0.1)}    & \textbf{99.6(0.1)}    & \textbf{99.6(0.1)} & \underline{98.9(0.3)}  & \textbf{99.6(0.1)}   & \textbf{99.7(0.1)}       \\
ECG5000                  & 93.6(0.7) & 93.0(0.2)  & 94.3(0.7)    & 94.4(0.2)    & \underline{95.3(0.2)} & 92.4(0.2)  & \textbf{95.6(0.3)}   & \underline{95.3(0.2)}       \\
TwoPatterns              & \underline{96.8(0.9)} & 42.2(10.3)  & \textbf{99.9(0.1)}    & \textbf{99.9(0.2)}    & \textbf{100.0(0.0)} & 84.2(2.5) & \textbf{100.0(0.0)}    & \textbf{100.0(0.0)}      \\
FordA                    & \underline{94.5(0.4)} & 93.3(0.7)  & {93.2(0.5)}    & 91.2(0.5)    & {93.2(0.4)} & 70.4(3.9)  & \textbf{95.0(0.4)}     & 92.2(0.6)       \\
UWaveGestureLibraryAll   & 79.0(0.7) & 83.0(1.4)  & 78.0(8.6)    & 85.5(1.4)    & \underline{89.2(0.4)} & 59.1(2.7)  & \textbf{94.5(0.7)}  & \underline{89.7(1.0)}       \\
FordB                    & \underline{93.1(0.5)} & 91.6(1.0)  & 91.0(0.5)    & 91.8(0.4)    & {92.1(0.5)}  &  63.6(7.4)  & \textbf{93.7(0.5)}    & 90.8(0.3)       \\
ChlorineConcentration    & 54.9(1.4) & 56.4(1.4)  & 85.4(1.1)    & 56.7(0.7)    & 81.4(0.9) & 56.6(0.4)  & \underline{88.4(0.9)}   & \textbf{93.2(1.1)}       \\
\hdashline
ShapesAll                  & \underline{73.6(2.9)} & 55.1(2.1) & 27.6(11.2)    & 48.8(3.5)   & 53.6(0.6) & 31.8(9.9)  & \textbf{79.6(1.9)}    & 66.6(3.6)       \\
FiftyWords                 & 44.0(0.8) & 29.1(1.2) & 31.7(1.3)    & 47.1(1.9)    & 46.7(2.7) & 25.0(2.1)  & \textbf{56.4(1.7)}     & \underline{49.0(1.9)}       \\
NonInvasiveFetalECGThorax1 & 32.3(2.9) & 50.3(13.2) & 72.9(3.2)    & 49.8(2.1)    & 79.7(1.1) & 40.7(5.3)  & \underline{83.0(0.8)}    & \textbf{87.4(0.7)}       \\
Phoneme                    & \underline{35.1(0.5)} & 34.7(2.2) & 27.5(2.0)    & 27.8(1.0)    & 34.5(1.1) & 23.7(4.0)  & \textbf{41.5(1.4)}    & 32.8(1.6)       \\
WordSynonyms               & 39.8(1.0) & 30.6(1.9) & 32.7(4.9)    & 42.4(2.7)    & \underline{47.7(3.3)} & 30.0(7.9)  & \textbf{55.9(2.9)}     & \underline{47.8(3.2)}      \\

\midrule
Mean Rank & 5.1	& 5.4	& 5.1	& 5.2	& 3.1	& 7.4	& 1.6	& 3.1 \\

\botrule
\end{tabular*}
\end{minipage}
\end{center}
\end{sidewaystable}

\begin{figure}[H]
    \centering
    \includegraphics[width=1.0\textwidth]{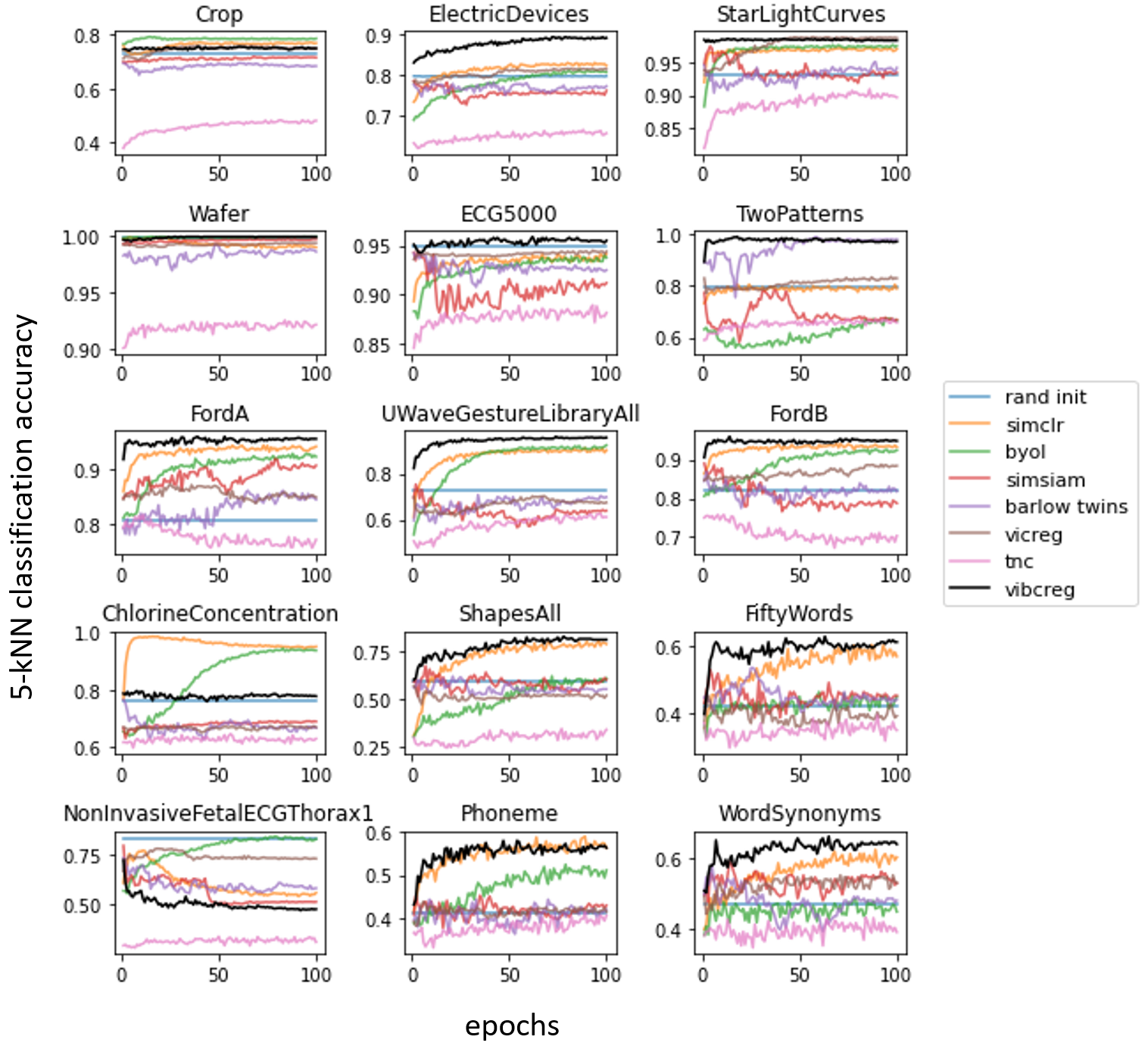}
    \caption{5-kNN classification accuracy on the UCR datasets during the representation learning.}
    \label{fig:UCR_knn_acc}
\end{figure}

\paragraph{Faster Representation Learning by VIbCReg}
\label{subsection:faster_representation_learning_by_vibcreg}
On top of the better performance, VIbCReg has another strength: \textit{faster representation learning}, which makes VIbCReg more appealing compared to its competitors. The speed of representation learning is presented by kNN classification accuracy for the UCR datasets \cite{Chen2020ExploringLearning} in Fig. \ref{fig:UCR_knn_acc} along with \textit{FD metric} $\mathcal{M}_{\mathrm{FD}}$ and \textit{FcE metric} $\mathcal{M}_{\mathrm{FcE}}$. The FD metric and FcE metric are metrics for the feature decorrleation and feature component expressiveness. The low FD metric indicates high feature decorrelation (\textit{i.e.,} features are well decorrelated) and vice versa. The low FcE metric indicates the feature collapse and vice versa. Note that VIbCReg is trained to have the FcE metric of 1. Details of both metrics are specified in Appendix \ref{appendix:FD_and_FcE_metrics}. To keep this section neat, the corresponding FD and FcE metrics of Fig. \ref{fig:UCR_knn_acc} is presented in Appendix \ref{appendix:faster_representation_learning_by_vibcreg}. It is recognizable that VIbCReg shows the fastest convergence and the highest kNN accuracy on most of the datasets.


\subsection{Between VICReg and VIbCReg}
\label{subsection:btn_vicreg_and_vibcreg}
\begin{figure}[H]
    \centering
    \includegraphics[width=\textwidth]{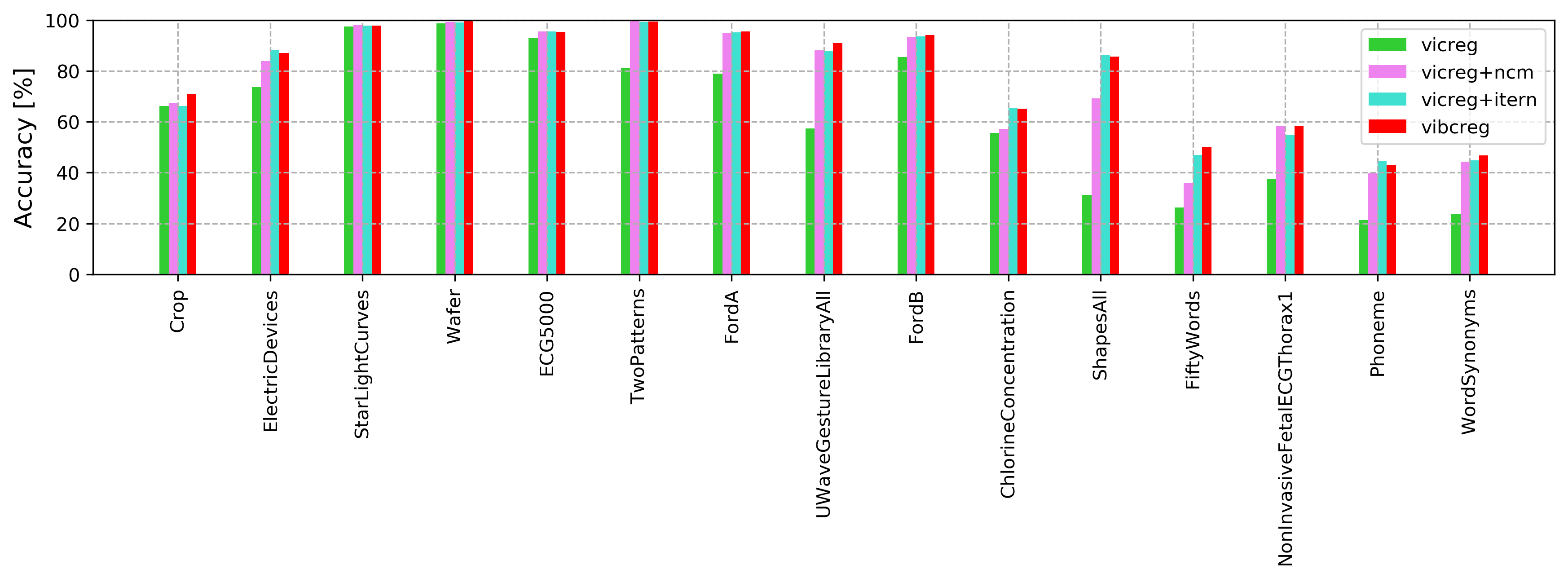}
    \caption{Comparative linear evaluation of the frameworks between VICReg and VIbCReg.}
    \label{fig:LE_btn_vicreg_vibcreg}
\end{figure}

As mentioned earlier in Section \ref{section:proposed_method}, VIbCReg can be viewed as VICReg with the normalized covariance matrix (NCM) and IterNorm. In this subsection, frameworks between VICReg and VIbCReg are investigated as shown in Table \ref{table:between_vicreg_and_vibcreg}. The linear evaluation results of the four frameworks on the UCR datasets are presented in Fig. \ref{fig:LE_btn_vicreg_vibcreg}. It shows that either adding  NCM (\texttt{ncm}) or IterNorm (\texttt{itern}) improves the performance. VIbCReg shows the most consistently-high performance among the four frameworks.

\begin{linespread}{1.0}
\begin{table}[ht]
\centering
\caption{Frameworks between VICReg and VIbCReg: VICReg+NCM and VICReg+IterN.}
\label{table:between_vicreg_and_vibcreg}
\small
\begin{tabular}{cccc}
\toprule
Frameworks & Normalized Covariance Matrix & IterNorm & Notation     \\
\midrule
VICReg     &                              &          & VICReg       \\
-          & o                            &          & \textit{VICReg+NCM}   \\
-          &                              & o        & \textit{VICReg+IterN} \\
-          & o                            & o        & VIbCReg      \\
\botrule
\end{tabular}
\end{table}
\end{linespread}

\subsection{Sensitivity Test w.r.t Weight for the FD Loss}
\begin{figure}[H]
    \centering
    \includegraphics[width=1.0\textwidth]{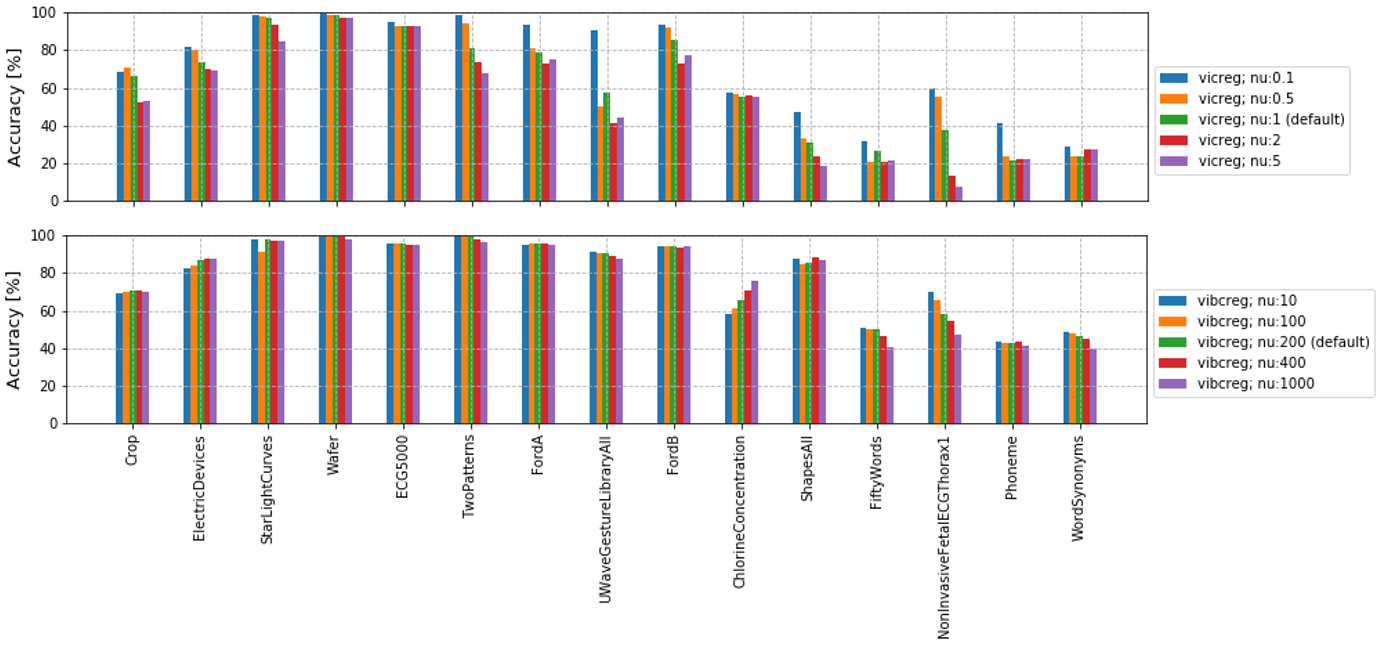}
    \caption{Linear evaluation of the sensitivity test with respect to $\nu$. Default values of $\nu$ for VICReg and VIbCReg are 1 and 200, respectively. Variants of $\nu$ are set by 5/10\%, 50\%, 200\%, and 500\% of the default values.}
    \label{fig:sensitivity_test_wrt_nu}
\end{figure}

The normalized covariance matrix is proposed to alleviate an effort for tuning the weight hyperparameter $\nu$ for the feature decorrelation loss term $c(Z)$. Without the normalization, a scale of $c(Z)$ from the covariance matrix can be significantly large and wide, which would make the tuning process harder. To show that the tuning process is easier with the normalized covariance matrix (\textit{i.e.,} performance is relatively quite consistent with respect to $\nu$), sensitivity test results with respect to $\nu$ are presented in Fig. \ref{fig:sensitivity_test_wrt_nu}. It is apparent that the performance gaps between different $\nu$ are much smaller for VIbCReg than VICReg in general, which makes the tuning process for $\nu$ much easier.

\subsection{Experimental Evaluation: Part 2}

\paragraph{Datasets}
The part-2 evaluation is conducted on the entire UCR and UEA archives except for a few datasets with varying length, significantly-short length such that an encoder cannot process, or missing values. The training and test datasets are used as given by the archives and dataset preprocessing follows the Franceschi which uses a simple z-normalization.

\paragraph{SVM evaluation}
An extensive experiment on the UCR and UEA archives was conducted by \cite{franceschi2019unsupervised} and the SVM evaluation is used in their work. We follow the same evaluation process as \cite{franceschi2019unsupervised} which is fitting a SVM-classifier on the learned representations with the penalty $C \in \{ 10^{i} \; \rvert \; i \in [-4, 4]\} \cup \{\infty\}$ and reporting the highest accuracy. 

\cite{franceschi2019unsupervised} reported several accuracies with respect to a different number of negative samples for each dataset. In our SVM evaluation result table, a mean of the accuracies is reported for the Franceschi. \cite{franceschi2019unsupervised} also reported SVM accuracy based on a concatenated representation of all the representations from different encoders with a different number of negative samples but that accuracy is not used in our comparison because that accuracy is based on multiple different encoders (\textit{i.e.,} 4 for UCR and 3 for UEA) that are trained separately. 

Complete result tables on UCR and UEA are presented in Tables \ref{tab:svm_acc_ucr}-\ref{tab:svm_acc_ucr_continued} and Table \ref{tab:svm_acc_uea}, respectively. There are several blank results on some UCR and UEA datasets for the computer-vision SSL framework and VIbCReg. That is due to the GPU memory issue caused by using two input pairs with crop length of 50\% and 100\% as described in Appendix \ref{appendix:data_augmentation_methods}.

It can be observed that VIbCReg performs better than all the other computer-vision SSL frameworks and performs better than the Franceschi on the UCR and similar to the Franceschi on the UEA. Yet, given that VIbCReg uses a simpler encoder, simpler SSL framework, and compatible with image data, VIbCReg is more appealing than the Franceschi overall. 
For the encoder, VIbCReg uses the common encoder, ResNet, while the Franceschi uses a specifically-modified TCN encoder. For the SSL framework, a non-contrastive approach is simpler than a contrastive one because it does not require negative example sampling. In the Franceschi, the performance can vary significantly by a choice of number of negative examples due to the nature of the contrastive learning. VIbCReg can be used not only for time series but also for image data unlike the Franceschi. Note that VIbCReg reaches the same performance as VICReg on images while VIbCReg performs better than VICReg on time series. VIbCReg for visual representation learning is available at \url{https://github.com/vturrisi/solo-learn} which includes results on CIFAR-10, CIFAR-100, and ImageNet-100 along with other popular computer-vision SSL frameworks. 

The performance ranking order remains the same in Table \ref{table:LE_UCR} and Table \ref{tab:svm_acc_ucr} for VIbCReg, SimCLR, VICReg, and Barlow Twins. But BYOL and SimSiam differ in their orders in the two tables. The main difference between the Part 1 and Part 2 experiments is the use of multi-crop with different crop sizes as introduced in Appendix \ref{appendix:data_augmentation_part2}. Therefore, we deduce that SimSiam has a good synergistic effect with the multi-crop while BYOL has the opposite effect.







\begin{linespread}{1.0}
\begin{table}[ht!]
    \centering
    \caption{Summary of the SVM accuracy on UCR.}
    \label{tab:svm_acc_ucr}
    \tiny
\begin{tabular}{l c c c c c c c c}
        \hline
        Subset names & \specialcell{Sim\\CLR} & BYOL & \specialcell{Sim\\Siam} & \specialcell{Barlow\\Twins} & VICReg & VIbCReg & \specialcell{TimeNet\\\cite{malhotra2017timenet}} & \specialcell{Frances.\\\cite{franceschi2019unsupervised}} \\ 
        \hline
        ACSF1 & 0.833 & 0.770 & 0.883 & 0.817 & 0.840 & 0.897 & ~ & 0.885 \\ 
        Adiac & 0.761 & 0.614 & 0.781 & 0.743 & 0.704 & 0.813 & 0.565 & 0.706 \\ 
        ArrowHead & 0.815 & 0.792 & 0.792 & 0.737 & 0.731 & 0.811 & ~ & 0.805 \\ 
        Beef & 0.489 & 0.678 & 0.711 & 0.511 & 0.667 & 0.733 & ~ & 0.667 \\ 
        BeetleFly & 0.667 & 0.650 & 0.783 & 0.717 & 0.700 & 0.783 & ~ & 0.850 \\ 
        BirdChicken & 0.950 & 0.900 & 0.933 & 0.900 & 0.933 & 0.933 & ~ & 0.813 \\ 
        BME & 0.889 & 0.904 & 0.984 & 0.751 & 0.802 & 0.971 & ~ & 0.991 \\ 
        Car & 0.778 & 0.717 & 0.789 & 0.594 & 0.711 & 0.861 & ~ & 0.746 \\ 
        CBF & 0.990 & 0.937 & 0.990 & 0.924 & 0.886 & 0.992 & ~ & 0.987 \\ 
        Chinatown & 0.958 & 0.921 & 0.972 & 0.956 & 0.914 & 0.954 & ~ & 0.949 \\ 
        ChlorineConcentration & 0.604 & 0.563 & 0.748 & 0.589 & 0.628 & 0.762 & 0.723 & 0.739 \\ 
        CinCECGtorso & 0.687 & 0.572 & 0.639 & 0.562 & 0.579 & 0.774 & ~ & 0.711 \\ 
        Coffee & 0.988 & 0.988 & 0.964 & 0.881 & 0.917 & 0.976 & ~ & 0.991 \\ 
        Computers & 0.711 & 0.653 & 0.777 & 0.669 & 0.689 & 0.785 & ~ & 0.668 \\ 
        CricketX & 0.704 & 0.574 & 0.785 & 0.553 & 0.624 & 0.809 & 0.659 & 0.715 \\ 
        CricketY & 0.729 & 0.544 & 0.767 & 0.562 & 0.621 & 0.811 & ~ & 0.692 \\ 
        CricketZ & 0.752 & 0.602 & 0.811 & 0.550 & 0.626 & 0.828 & ~ & 0.716 \\ 
        Crop & 0.720 & 0.700 & 0.740 & 0.676 & 0.759 & 0.744 & ~ & 0.726 \\ 
        DiatomSizeReduction & 0.877 & 0.886 & 0.919 & 0.892 & 0.859 & 0.859 & ~ & 0.988 \\ 
        DistalPhalanxOutlineAgeGroup & 0.741 & 0.736 & 0.717 & 0.727 & 0.727 & 0.743 & ~ & 0.728 \\ 
        DistalPhalanxOutlineCorrect & 0.774 & 0.756 & 0.772 & 0.737 & 0.761 & 0.767 & ~ & 0.764 \\ 
        DistalPhalanxTW & 0.662 & 0.655 & 0.650 & 0.671 & 0.662 & 0.652 & ~ & 0.678 \\ 
        Earthquakes & 0.748 & 0.748 & 0.748 & 0.748 & 0.748 & 0.748 & ~ & 0.748 \\ 
        ECG200 & 0.863 & 0.867 & 0.853 & 0.820 & 0.873 & 0.850 & ~ & 0.893 \\ 
        ECG5000 & 0.940 & 0.934 & 0.931 & 0.930 & 0.937 & 0.935 & 0.934 & 0.937 \\ 
        ECGFiveDays & 0.890 & 0.833 & 0.759 & 0.810 & 0.816 & 0.768 & ~ & 1.000 \\ 
        ElectricDevices & 0.681 & 0.687 & 0.697 & 0.636 & 0.598 & 0.682 & 0.665 & 0.707 \\ 
        EOGHorizontalSignal & 0.527 & 0.341 & 0.555 & 0.376 & 0.432 & 0.645 & ~ & 0.565 \\ 
        EOGVerticalSignal & 0.366 & 0.283 & 0.373 & 0.320 & 0.311 & 0.405 & ~ & 0.419 \\ 
        EthanolLevel & 0.416 & 0.432 & 0.301 & 0.439 & 0.550 & 0.308 & ~ & 0.364 \\ 
        FaceAll & 0.740 & 0.699 & 0.760 & 0.676 & 0.712 & 0.800 & ~ & 0.773 \\ 
        FaceFour & 0.856 & 0.720 & 0.909 & 0.894 & 0.814 & 0.924 & ~ & 0.847 \\ 
        FacesUCR & 0.821 & 0.753 & 0.867 & 0.733 & 0.795 & 0.891 & ~ & 0.882 \\ 
        FiftyWords & 0.673 & 0.500 & 0.709 & 0.560 & 0.565 & 0.759 & ~ & 0.739 \\ 
        Fish & 0.865 & 0.832 & 0.937 & 0.775 & 0.785 & 0.981 & ~ & 0.899 \\ 
        FordA & 0.949 & 0.655 & 0.956 & 0.877 & 0.923 & 0.952 & ~ & 0.925 \\ 
        FordB & 0.854 & 0.826 & 0.848 & 0.728 & 0.814 & 0.846 & ~ & 0.787 \\ 
        FreezerRegularTrain & 0.931 & 0.979 & 0.967 & 0.950 & 0.971 & 0.963 & ~ & 0.977 \\ 
        FreezerSmallTrain & 0.744 & 0.910 & 0.762 & 0.762 & 0.725 & 0.877 & ~ & 0.941 \\ 
        Fungi & ~ & ~ & ~ & ~ & ~ & ~ & ~ & 1.000 \\ 
        GunPoint & 0.944 & 0.911 & 0.978 & 0.938 & 0.962 & 0.987 & ~ & 0.977 \\ 
        GunPointAgeSpan & 0.960 & 0.961 & 0.966 & 0.977 & 0.957 & 0.980 & ~ & 0.989 \\ 
        GunPointMaleVersusFemale & 0.997 & 0.996 & 0.997 & 0.996 & 0.988 & 0.999 & ~ & 0.999 \\ 
        GunPointOldVersusYoung & 1.000 & 0.990 & 0.990 & 1.000 & 0.999 & 0.995 & ~ & 1.000 \\ 
        Ham & 0.613 & 0.619 & 0.622 & 0.546 & 0.625 & 0.635 & ~ & 0.679 \\ 
        HandOutlines & 0.894 & 0.796 & 0.886 & 0.753 & 0.876 & 0.907 & ~ & 0.918 \\ 
        Haptics & 0.468 & 0.367 & 0.461 & 0.432 & 0.405 & 0.504 & ~ & 0.451 \\ 
        Herring & 0.589 & 0.578 & 0.557 & 0.589 & 0.568 & 0.578 & ~ & 0.594 \\ 
        HouseTwenty & 0.927 & 0.734 & 0.964 & 0.826 & 0.832 & 0.975 & ~ & 0.933 \\ 
        InlineSkate & 0.353 & 0.310 & 0.406 & 0.358 & 0.286 & 0.446 & ~ & 0.413 \\ 
        InsectEPGRegularTrain & 1.000 & 1.000 & 1.000 & 1.000 & 1.000 & 0.999 & ~ & 1.000 \\ 
        InsectEPGSmallTrain & 1.000 & 0.941 & 1.000 & 1.000 & 0.989 & 0.996 & ~ & 1.000 \\ 
        InsectWingbeatSound & 0.546 & 0.485 & 0.539 & 0.474 & 0.506 & 0.566 & ~ & 0.604 \\ 
        ItalyPowerDemand & 0.951 & 0.926 & 0.936 & 0.898 & 0.928 & 0.942 & ~ & 0.937 \\ 
        LargeKitchenAppliances & 0.848 & 0.752 & 0.875 & 0.761 & 0.739 & 0.876 & ~ & 0.814 \\ 
        Lightning2 & 0.732 & 0.738 & 0.732 & 0.650 & 0.656 & 0.749 & ~ & 0.857 \\ 
        Lightning7 & 0.653 & 0.571 & 0.667 & 0.598 & 0.708 & 0.689 & ~ & 0.809 \\ 
        Mallat & 0.942 & 0.861 & 0.889 & 0.866 & 0.903 & 0.871 & ~ & 0.948 \\ 
        Meat & 0.894 & 0.950 & 0.906 & 0.911 & 0.928 & 0.861 & ~ & 0.900 \\ 
        MedicalImages & 0.749 & 0.700 & 0.760 & 0.674 & 0.726 & 0.764 & 0.753 & 0.757 \\ 
        MelbournePedestrian & 0.921 & 0.903 & 0.906 & 0.861 & 0.930 & 0.918 & ~ & 0.946 \\ 
        MiddlePhalanxOutlineAgeGroup & 0.565 & 0.630 & 0.606 & 0.545 & 0.580 & 0.571 & ~ & 0.648 \\ 
        MiddlePhalanxOutlineCorrect & 0.694 & 0.758 & 0.806 & 0.788 & 0.821 & 0.780 & ~ & 0.792 \\ 
        MiddlePhalanxTW & 0.604 & 0.530 & 0.593 & 0.582 & 0.584 & 0.593 & ~ & 0.604 \\ 
        MixedShapesRegularTrain & 0.945 & 0.865 & 0.961 & 0.928 & 0.876 & 0.962 & ~ & 0.908 \\ 
        MixedShapesSmallTrain & 0.872 & 0.727 & 0.885 & 0.805 & 0.748 & 0.905 & ~ & 0.863 \\ 
        MoteStrain & 0.815 & 0.789 & 0.877 & 0.747 & 0.820 & 0.871 & ~ & 0.862 \\ 
        NonInvasiveFetalECGThorax1 & 0.809 & 0.758 & 0.777 & 0.834 & 0.886 & 0.832 & ~ & 0.901 \\ 
        NonInvasiveFetalECGThorax2 & 0.881 & 0.837 & 0.857 & 0.879 & 0.903 & 0.910 & ~ & 0.922 \\ 
        OliveOil & 0.733 & 0.844 & 0.744 & 0.811 & 0.800 & 0.811 & ~ & 0.859 \\ 
        OSULeaf & 0.851 & 0.642 & 0.868 & 0.726 & 0.679 & 0.895 & ~ & 0.742 \\ 
        PhalangesOutlinesCorrect & 0.746 & 0.625 & 0.797 & 0.646 & 0.737 & 0.818 & 0.772 & 0.796 \\ 
        Phoneme & 0.292 & 0.201 & 0.315 & 0.215 & 0.239 & 0.325 & ~ & 0.264 \\ 
        PigAirwayPressure & ~ & ~ & ~ & ~ & ~ & ~ & ~ & 0.464 \\ 
        PigArtPressure & ~ & ~ & ~ & ~ & ~ & ~ & ~ & 0.921 \\ 
        PigCVP & ~ & ~ & ~ & ~ & ~ & ~ & ~ & 0.590 \\ 
        Plane & 0.987 & 0.971 & 0.978 & 0.975 & 0.978 & 0.990 & ~ & 0.992 \\ 
        PowerCons & 0.970 & 0.931 & 0.957 & 0.957 & 0.974 & 0.980 & ~ & 0.925 \\ 
        ProximalPhalanxOutlineAgeGroup & 0.839 & 0.834 & 0.850 & 0.826 & 0.837 & 0.844 & ~ & 0.849 \\ 
        ProximalPhalanxOutlineCorrect & 0.795 & 0.794 & 0.834 & 0.795 & 0.847 & 0.866 & ~ & 0.862 \\ 
        ProximalPhalanxTW & 0.772 & 0.771 & 0.802 & 0.797 & 0.795 & 0.789 & ~ & 0.793 \\ 
        RefrigerationDevices & 0.500 & 0.500 & 0.516 & 0.498 & 0.528 & 0.531 & ~ & 0.525 \\ 
        Rock & 0.440 & 0.447 & 0.447 & 0.453 & 0.400 & 0.687 & ~ & 0.600 \\ 
        ScreenType & 0.517 & 0.421 & 0.524 & 0.434 & 0.460 & 0.561 & ~ & 0.416 \\ 
        SemgHandGenderCh2 & 0.877 & 0.730 & 0.877 & 0.849 & 0.880 & 0.847 & ~ & 0.865 \\ 
        \hline
    \end{tabular}
\end{table}
\end{linespread}

\begin{linespread}{1.0}
\begin{table}[ht!]
    \centering
    \caption{Summary of the SVM accuracy on UCR (continued).}
    \label{tab:svm_acc_ucr_continued}
    \tiny
\begin{tabular}{l c c c c c c c c}
        \hline
        Subset names & \specialcell{Sim\\CLR} & BYOL & \specialcell{Sim\\Siam} & \specialcell{Barlow\\Twins} & VICReg & VIbCReg & \specialcell{TimeNet\\\cite{malhotra2017timenet}} & \specialcell{Frances.\\\cite{franceschi2019unsupervised}} \\ 
        \hline
        SemgHandMovementCh2 & 0.779 & 0.596 & 0.739 & 0.650 & 0.684 & 0.676 & ~ & 0.712 \\ 
        SemgHandSubjectCh2 & 0.842 & 0.666 & 0.821 & 0.768 & 0.808 & 0.805 & ~ & 0.822 \\ 
        ShapeletSim & 0.720 & 0.622 & 0.767 & 0.857 & 0.574 & 0.757 & ~ & 0.674 \\ 
        ShapesAll & 0.856 & 0.696 & 0.882 & 0.751 & 0.776 & 0.910 & ~ & 0.848 \\ 
        SmallKitchenAppliances & 0.795 & 0.696 & 0.837 & 0.721 & 0.728 & 0.834 & ~ & 0.685 \\ 
        SmoothSubspace & 0.813 & 0.847 & 0.776 & 0.820 & 0.853 & 0.862 & ~ & 0.948 \\ 
        SonyAIBORobotSurface1 & 0.784 & 0.722 & 0.733 & 0.772 & 0.708 & 0.818 & ~ & 0.893 \\ 
        SonyAIBORobotSurface2 & 0.884 & 0.730 & 0.923 & 0.757 & 0.832 & 0.860 & ~ & 0.909 \\ 
        StarLightCurves & 0.972 & 0.936 & 0.978 & 0.961 & 0.960 & 0.977 & ~ & 0.962 \\ 
        Strawberry & 0.957 & 0.939 & 0.954 & 0.932 & 0.959 & 0.963 & ~ & 0.951 \\ 
        SwedishLeaf & 0.934 & 0.835 & 0.947 & 0.902 & 0.893 & 0.961 & 0.901 & 0.918 \\ 
        Symbols & 0.928 & 0.731 & 0.960 & 0.891 & 0.896 & 0.975 & ~ & 0.949 \\ 
        SyntheticControl & 0.983 & 0.958 & 0.978 & 0.939 & 0.959 & 0.979 & 0.983 & 0.984 \\ 
        ToeSegmentation1 & 0.942 & 0.789 & 0.953 & 0.655 & 0.731 & 0.936 & ~ & 0.925 \\ 
        ToeSegmentation2 & 0.815 & 0.738 & 0.900 & 0.662 & 0.731 & 0.903 & ~ & 0.875 \\ 
        Trace & 0.987 & 0.920 & 1.000 & 0.993 & 0.963 & 0.997 & ~ & 0.998 \\ 
        TwoLeadECG & 0.762 & 0.706 & 0.934 & 0.797 & 0.713 & 0.911 & 0.999 & 0.996 \\ 
        TwoPatterns & 0.990 & 0.997 & 0.975 & 0.675 & 0.906 & 0.993 & ~ & 0.999 \\ 
        UMD & 0.877 & 0.833 & 0.977 & 0.875 & 0.852 & 0.979 & ~ & 0.988 \\ 
        UWaveGestureLibraryAll & 0.924 & 0.730 & 0.883 & 0.780 & 0.790 & 0.918 & ~ & 0.895 \\ 
        UWaveGestureLibraryX & 0.802 & 0.715 & 0.795 & 0.677 & 0.741 & 0.801 & ~ & 0.794 \\ 
        UWaveGestureLibraryY & 0.747 & 0.631 & 0.749 & 0.598 & 0.649 & 0.757 & ~ & 0.711 \\ 
        UWaveGestureLibraryZ & 0.755 & 0.662 & 0.755 & 0.621 & 0.706 & 0.764 & ~ & 0.743 \\ 
        Wafer & 0.991 & 0.988 & 0.991 & 0.983 & 0.992 & 0.991 & 0.994 & 0.993 \\ 
        Wine & 0.772 & 0.722 & 0.765 & 0.617 & 0.772 & 0.846 & ~ & 0.797 \\ 
        WordSynonyms & 0.496 & 0.428 & 0.532 & 0.427 & 0.511 & 0.678 & ~ & 0.661 \\ 
        Worms & 0.758 & 0.606 & 0.792 & 0.610 & 0.684 & 0.788 & ~ & 0.704 \\ 
        WormsTwoClass & 0.827 & 0.706 & 0.831 & 0.688 & 0.719 & 0.827 & ~ & 0.763 \\ 
        Yoga & 0.849 & 0.776 & 0.884 & 0.759 & 0.817 & 0.896 & 0.866 & 0.837 \\ 
        \hline
        Mean Rank & 3.6 & 5.7 & 3.2 & 5.5 & 4.7 & 2.5 & ~ & 2.8 \\ 
        \hline
    \end{tabular}
\end{table}
\end{linespread}

\begin{linespread}{1.0}
\begin{table}[ht]
    \centering
    \caption{Summary of the SVM accuracy on UEA.}
    \label{tab:svm_acc_uea}
    \tiny
    \begin{tabular}{l c c c c c c c}
        \hline
        Subset names & \specialcell{Sim\\CLR} & BYOL & \specialcell{Sim\\Siam} & \specialcell{Barlow\\Twins} & VICReg & VIbCReg & \specialcell{Frances.\\\cite{franceschi2019unsupervised}} \\
        \hline
        ArticularyWordRecognition & 0.934 & 0.883 & 0.861 & 0.744 & 0.889 & 0.956 & 0.961 \\ 
        AtrialFibrillation & 0.311 & 0.400 & 0.289 & 0.267 & 0.378 & 0.311 & 0.133 \\ 
        BasicMotions & 0.983 & 0.850 & 0.983 & 0.992 & 0.900 & 1.000 & 1.000 \\ 
        Cricket & 0.963 & 0.884 & 0.968 & 0.931 & 0.944 & 0.968 & 0.967 \\ 
        DuckDuckGeese & 0.393 & 0.313 & 0.473 & 0.307 & 0.360 & 0.513 & 0.642 \\ 
        EigenWorms & ~ & ~ & ~ & ~ & ~ & ~ & 0.837 \\ 
        Epilepsy & 0.961 & 0.829 & 0.964 & 0.915 & 0.942 & 0.973 & 0.971 \\ 
        ERing & 0.862 & 0.848 & 0.844 & 0.798 & 0.859 & 0.901 & 0.133 \\ 
        EthanolConcentration & 0.294 & 0.307 & 0.319 & 0.270 & 0.279 & 0.304 & 0.248 \\ 
        FaceDetection & ~ & ~ & ~ & ~ & ~ & ~ & 0.520 \\ 
        FingerMovements & 0.513 & 0.497 & 0.527 & 0.540 & 0.530 & 0.520 & 0.540 \\ 
        HandMovementDirection & 0.311 & 0.315 & 0.356 & 0.275 & 0.297 & 0.360 & 0.320 \\ 
        Handwriting & 0.325 & 0.319 & 0.433 & 0.188 & 0.262 & 0.472 & 0.454 \\ 
        Heartbeat & 0.711 & 0.722 & 0.720 & 0.719 & 0.717 & 0.750 & 0.743 \\ 
        Libras & 0.850 & 0.667 & 0.857 & 0.735 & 0.748 & 0.863 & 0.881 \\ 
        LSST & 0.515 & 0.412 & 0.557 & 0.389 & 0.482 & 0.577 & 0.532 \\ 
        MotorImagery & 0.450 & 0.513 & 0.540 & 0.573 & 0.520 & 0.493 & 0.550 \\ 
        NATOPS & 0.802 & 0.678 & 0.759 & 0.750 & 0.724 & 0.839 & 0.922 \\ 
        PEMS-SF & 0.809 & 0.798 & 0.726 & 0.842 & 0.829 & 0.682 & 0.661 \\ 
        PenDigits & ~ & ~ & ~ & ~ & ~ & ~ & 0.982 \\ 
        Phoneme & 0.302 & 0.203 & 0.332 & 0.250 & 0.229 & 0.315 & 0.217 \\ 
        RacketSports & 0.789 & 0.739 & 0.803 & 0.671 & 0.728 & 0.789 & 0.822 \\ 
        SelfRegulationSCP1 & 0.716 & 0.755 & 0.732 & 0.784 & 0.779 & 0.724 & 0.821 \\ 
        SelfRegulationSCP2 & 0.552 & 0.496 & 0.506 & 0.539 & 0.546 & 0.513 & 0.543 \\ 
        StandWalkJump & ~ & ~ & ~ & ~ & ~ & ~ & 0.355 \\ 
        UWaveGestureLibrary & 0.817 & 0.685 & 0.801 & 0.591 & 0.625 & 0.861 & 0.876 \\ 
        \hline
        Mean Rank & 4.1 & 5.2 & 3.5 & 5.1 & 4.5 & 2.7 & 2.9 \\ 
        \hline
    \end{tabular}
\end{table}
\end{linespread}

\section{Limitation}

VIbCReg is purposefully designed to stay similar to the computer-vision SSL frameworks so that it can be compatible with both images and time series. 
However, the current version of VIbCReg cannot process a very-long sequence due to a GPU memory limit since it employs the multi-crop where one crop with 50\% length and another crop with 100\% length of input time series. Also, the ResNet encoder is used for VIbCReg to comply with the computer-vision SSL frameworks. However, ResNet has a fixed receptive field size and can only process local information for long time series. Better performance is expected when an encoder that can enable various-sized receptive fields is used.

\section{Conclusion}
In this paper, we have evaluated the well-known Siamese-style computer vision SSL frameworks. The results show that even when the computer vision SSL frameworks are naively applied to a different modality (\textit{i.e.,} time series), they can still result in effective representation learning. We argue that some of the design ideas from the computer vision SSL frameworks may be taken to develop a better representation learning method for time series in future work. We also introduce VIbCReg and it shows the robust performance while outperforming the others. The good performance of VIbCReg is achieved by inducing stronger feature decorrelation with a normalized covariance and iterative normalization layer.


\bmhead{Acknowledgments}
We would like to thank the Norwegian Research Council for funding the Machine Learning for Irregular Time Series (ML4ITS) project (312062). This funding directly supported this research.

\section{Statements and Declarations}
\paragraph{Funding} This research is funded by the Norwegian Research Council for funding the Machine Learning for Irregular Time Series (ML4ITS) project (312062).
\paragraph{Conflicts of interest/Competing interests} Not applicable.
\paragraph{Ethics approval} Not applicable.
\paragraph{Consent to participate} Not applicable.
\paragraph{Consent for publication} Not applicable.
\paragraph{Availability of data and materials} The used datasets are popular open-source bench mark datasets and the used deep learning library is an open-source Python deep learning library named PyTorch.

\paragraph{Authors’ Contributions}

\begin{enumerate}
    \item Daesoo Lee: Proposed a first draft of the proposed method and improved the proposal and implemented the proposal with his supervisor, Prof. Erlend Aune. And Daesoo worked on the paper with Prof. Aune.
    \item Erlend Aune: Improved the initial proposal and implemented the proposal and wrote the paper with Daesoo.
\end{enumerate}



\begin{appendices}


\section{Implementation Details}

\begin{table}[ht]
\footnotesize
\centering
\caption{PyTorch-style pseudocode for VIbCReg. We mostly follow the same notations from the VICReg paper.}
\label{tab:pseudocode_vibcreg}
\begin{tabular}{l}
\toprule
\textbf{Algorithm 1} PyTorch-style pseudocode for VIbCReg \\
\midrule
\begin{lstlisting}
# f: encoder network
# lambda_, mu, nu: coefficients of the invariance, variance,  
#                  and covariance losses
# N: batch size
# F: feature size (= dimension of the representations)
# 
# mse_loss: Mean square error loss function
# off_diagonal: off-diagonal elements of a matrix
# relu: ReLU activation function
# normalize: torch.nn.functional.normalize(..)

for x in loader: # load a batch with N samples
    # two randomly augmented versions of x
    x_a, x_b = augment(x)
    
    # compute representations
    z_a = f(x_a)  # N x F
    z_b = f(x_b)  # N x F
    
    # invariance loss
    sim_loss = mse_loss(z_a, z_b)
    
    # variance loss
    std_z_a = torch.sqrt(z_a.var(dim=0) + 1e-4)
    std_z_b = torch.sqrt(z_b.var(dim=0) + 1e-4)
    std_loss = torch.mean(relu(1 - std_z_a))
    std_loss = std_loss + torch.mean(relu(1 - std_z_b))
    
    # covariance loss
    z_a = z_a - z_a.mean(dim=0)
    z_b = z_b - z_b.mean(dim=0)
    norm_z_a = normalize(z_a, p=2, dim=0)
    norm_z_b = normalize(z_b, p=2, dim=0)
    norm_cov_z_a = (norm_z_a.T @ norm_z_a)
    norm_cov_z_b = (norm_z_b.T @ norm_z_b)
    norm_cov_loss = (off_diagonal(norm_cov_z_a)**2).mean() \
                    + (off_diagonal(norm_cov_z_b)**2).mean()
    
    # loss
    loss = lambda_ * sim_loss + mu * std_loss + nu * norm_cov_loss
    
    # optimization step
    loss.backward()
    optimizer.step()
    
\end{lstlisting} \\[0.7cm]

\botrule 

\end{tabular}
\end{table}

\subsection{Encoder}
\begin{figure}[H]
    \centering
    \includegraphics[height=12cm]{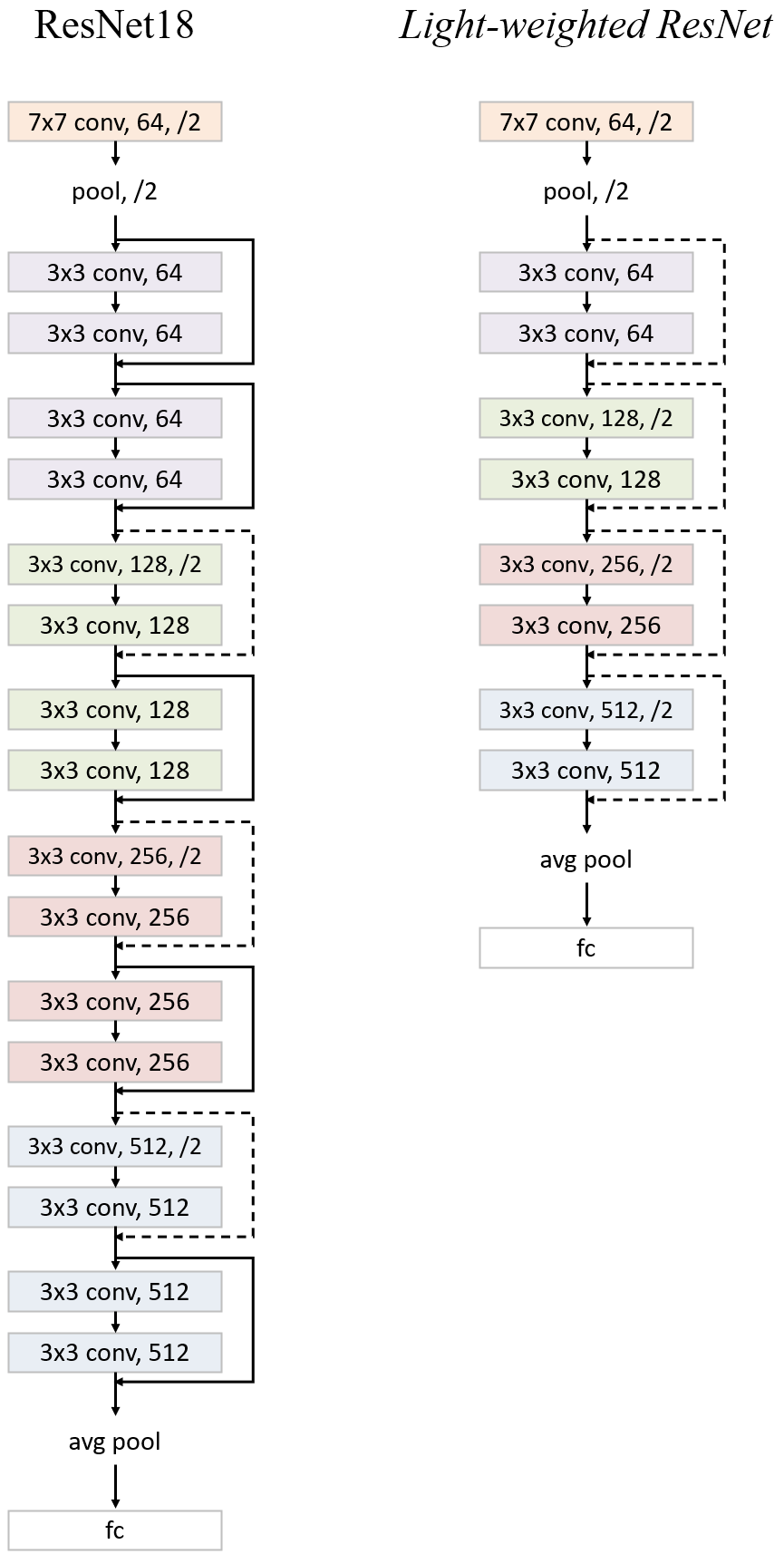}
    \caption{The light-weighted ResNet is illustrated compared to ResNet18.}
    \label{fig:resnet}
\end{figure}

\subsection{SSL Frameworks}

In our experiments, the SSL frameworks are implemented as close as possible to the original implementations. The only major difference is the encoder's dimension. Instead of 2-dimensional image input, they receive 1-dimensional time series input. Unless specified differently, the architectures follow original implementations.

\paragraph{SimCLR} Projector's dimension size is set to 4096. $\tau$ is set to 0.1.

\paragraph{BYOL} Projector's dimension size is set to 512. The momentum of 0.9 is used. 

\paragraph{SimSiam} All the settings are the same as its original implementation.

\paragraph{Barlow Twins} Projector's dimension size is set to 4096. $\lambda$ is set to $5\cdot10^{-3}$.

\paragraph{VICReg} Projector's dimension size is set to 4096. $\lambda$, $\mu$, and $\nu$ are set to 25, 25, and 1, respectively.

\paragraph{VIbCReg} Its projector consists as follows: (Linear-BN-ReLU)-(Linear-BN-ReLU)-(Linear-IterNorm), where Linear, BN, and ReLU denotes a linear layer, batch normalization \cite{Ioffe2015BatchShift}, and rectified linear unit, respectively. The dimension of the inner and output layers of the projector is set to 4096. $\lambda$ and $\mu$ are set to 25 and 25, respectively. $\nu$ is set to 200 for the part-1 evaluation and 100 for the part-2 evaluation.


\paragraph{TNC} Its discriminator consists as follows: Linear-ReLU-Dropout(0.5)-Linear, following its original implementation in GitHub (\url{https://github.com/sanatonek/TNC_representation_learning}). The discriminator's input dimension size is 2 times the representation size and its hidden layer's dimension size is 4 times the representation size. $w$ is set to 0.05. As for selecting its positive and negative examples, a time step for a reference example is first randomly sampled. Note that the reference example is a random crop with the sampled time step at the center. Then, a time step for a positive example is sampled according to a normal distribution with mean of the time step of the reference example and standard deviation of crop length/4. A time step for a negative example is sampled outside of the normal distribution for the positive example.

\subsection{Data Augmentation Methods}
\label{appendix:data_augmentation_methods}

\paragraph{Data Augmentation for Experimental Evaluation Part 1}
\label{appendix:data_augmentation_part1}
In the part-1 evaluation, three data augmentation methods are used for training: \texttt{Random Crop}, \texttt{Random Amplitude Resize}, and \texttt{Random Vertical Shift}, and \texttt{Random Amplitude Resize} and \texttt{Random Vertical Shift} are used for the linear evaluation.

\texttt{Random Crop} is similar to the one in computer vision. The only difference is that the random crop is conducted on 1D time series data in our experiments. Its hyperparameter is crop size. The crop size for the UCR datasets in the part-1 evaluation is presented in Table \ref{tab:summary_crop_size}. 
Different datasets have different patterns to be distinguishable with respect to different classes. Some have long patterns while some have short patterns. Depending on length of distinguishable patterns, ideal crop size may differ. For the crop length in the table, the ones that result in better performance are chosen. A more generic approach is used in the part-2 evaluation.

\begin{table}[ht]
\centering
\caption{Summary of the crop size for each UCR dataset.}
\label{tab:summary_crop_size}
\small
\begin{tabular}{l c c c}
\toprule
Dataset name               & Length & \specialcell{Use half length \\ for crop} & Crop size \\
\midrule
Crop                       & 46   & x & 46  \\
ElectricDevices            & 96   & o & 48  \\
StarLightCurves            & 1024 & o & 512 \\
Wafer                      & 152  & x & 152 \\
ECG5000                    & 140  & o & 70  \\
TwoPatterns                & 128  & o & 64  \\
FordA                      & 500  & o & 250 \\
UWaveGestureLibraryAll     & 945  & o & 473 \\
FordB                      & 500  & o & 250 \\
ChlorineConcentration      & 166  & x & 166 \\
ShapesAll                  & 512  & o & 256 \\
FiftyWords                 & 270  & o & 135 \\
NonInvasiveFetalECGThorax1 & 750  & x & 750 \\
Phoneme                    & 1024 & o & 512 \\
WordSynonyms               & 270  & o & 135 \\
\botrule
\end{tabular}
\end{table}

\texttt{Random Amplitude Resize} randomly resizes overall amplitude of input time series. It is expressed as Eq. (\ref{eq:random_amplitude_resize}), where $m_{\mathrm{rar}}$ is a multiplier to input time series $x$, and $\alpha_{\mathrm{rar}}$ is a hyperparameter. $\alpha_{\mathrm{rar}}$ is set to 0.3.

\begin{equation}
    x = m_{\mathrm{rar}} \: x; \;\;\; m_{\mathrm{rar}} \sim \mathrm{U}(1-\alpha_{\mathrm{rar}}, 1+\alpha_{\mathrm{rar}})
    \label{eq:random_amplitude_resize}
\end{equation}

\texttt{Random Vertical Shift} randomly shifts input time series in a vertical direction. It is expressed as Eq. (\ref{eq:random_vertical_shift}), where $x^\prime$ denotes input time series before any augmentation, $\mathrm{Std}$ denotes a standard deviation estimator, and $\beta_{\mathrm{rvs}}$ is a hyperparameter to determine a magnitude of the vertical shift. $\beta_{\mathrm{rvs}}$ is set to 0.5. 
\begin{equation}
    x = x + s_{\mathrm{rvs}}; \;\;\; s_{\mathrm{rvs}} \sim \mathrm{U}(-\alpha_{\mathrm{rvs}}, \alpha_{\mathrm{rvs}})
    \label{eq:random_vertical_shift}
\end{equation}
\begin{equation}
    \alpha_{\mathrm{rvs}} = \beta_{\mathrm{rvs}} \: \mathrm{Std}(x^\prime)
\end{equation}

\paragraph{Data Augmentation for Experimental Evaluation Part 2}
\label{appendix:data_augmentation_part2}
In the part-2 evaluation, two data augmentation methods are used: \texttt{Random Crop} and \texttt{Random Amplitude Resize}.

\paragraph{Random Crop} Instead of using the better crop length between 50\% and 100\% of time series length, both are used in the part-2 evaluation such that the invariance, variance, and the covariance loss are computed twice for two different input pairs with different crop lengths to allow different views for the models. It is inspired by \cite{caron2020unsupervised} that proposed SwAV, and they experimentally show that using multiple crops with different sizes helps to produce better representations. Then, the loss function is computed as in Table \ref{tab:loss_with_two_crop_lengths}.

\begin{table}[ht]
\footnotesize
\centering
\caption{Pseudocode for VIbCReg with Random Crop with two different crop lengths. The notation follows the pseudocode for VIbCReg in Table \ref{tab:pseudocode_vibcreg}. Note that \texttt{crop\_length\_ratio} represents different crop lengths: 0.5 and 1.0 denote 50\% and 100\%, respectively.}
\label{tab:loss_with_two_crop_lengths}
\begin{tabular}{l}
\toprule
\textbf{Algorithm 2} PyTorch-style pseudocode for VIbCReg with \texttt{Random Crop} with two different \\crop lengths \\
\midrule
\begin{lstlisting}
for x in loader: # load a batch with N samples
    for crop_length_ratio in [0.5, 1.0]:
        # two randomly augmented versions of x 
        # with crop length of 50% or 100% of time series.
        x_a, x_b = augment(x, crop_length_ratio)
        
        # compute representations
        z_a = f(x_a)  # N x F
        z_b = f(x_b)  # N x F
        
        # loss
        invar_loss = variance_loss(z_a, z_b)
        var_loss = variance_loss(z_a, z_b)
        cov_loss = covariance_loss(z_a, z_b)
        loss = lambda_ * invar_loss + mu * var_loss + nu * cov_loss
        
        # optimization step
        loss.backward()
        optimizer.step()
    
\end{lstlisting} \\[0.7cm]
\botrule 
\end{tabular}
\end{table}

\texttt{Random Amplitude Resize} is slightly different in the part-2 evaluation from the part-1 evaluation as Eq. (\ref{eq:random_amplitude_resize_part2}). $\alpha_{\mathrm{rar}}$ is set to 0.1.

\begin{equation}
    x = m_{\mathrm{rar}} \: x; \;\;\; m_{\mathrm{rar}} \sim N(0, \alpha_{\mathrm{rar}})
    \label{eq:random_amplitude_resize_part2}
\end{equation}

\section{FD and FcE Metrics}
\label{appendix:FD_and_FcE_metrics}
FD and FcE metrics are metrics for the feature decorrleation and feature component expressiveness. They are used to keep track of FD and FcE status of learned features during SSL for representation learning. The FD metric and FcE metric are defined in Eq. (\ref{eq:FD_metrics}) and Eq. (\ref{eq:FcE_metrics}), respectively. $Z$ is output from the projector and $F$ is feature size of $Z$. $\sum_{i \neq j}$ denotes ignoring the diagonal terms in the matrix. In Eq. (\ref{eq:FD_metrics_CZ}), the l2-norm is conducted along the batch dimension. In Eq. (\ref{eq:FcE_metrics}), $\mathrm{Std}$ denotes a standard deviation estimator and it is conducted along the batch dimension.

\begin{equation}
    C(Z) = \left( \frac{Z - \bar{Z}}{ \| Z - \bar{Z} \|_2 } \right)^T \left( \frac{Z - \bar{Z}}{ \| Z - \bar{Z} \|_2 } \right)
    \label{eq:FD_metrics_CZ}
\end{equation}

\begin{equation}
    \mathcal{M}_{\mathrm{FD}} = \frac{1}{F^2} \sum_{i \neq j}{\abs{C(Z)}}
    \label{eq:FD_metrics}
\end{equation}

\begin{equation}
    \mathcal{M}_{\mathrm{FcE}} = \frac{1}{F} \sum^F_{f=1} \mathrm{Std}(Z)
    \label{eq:FcE_metrics}
\end{equation}

\section{Additional Result Materials}

\begin{figure}[H]
    \centering
    \includegraphics[width=\textwidth]{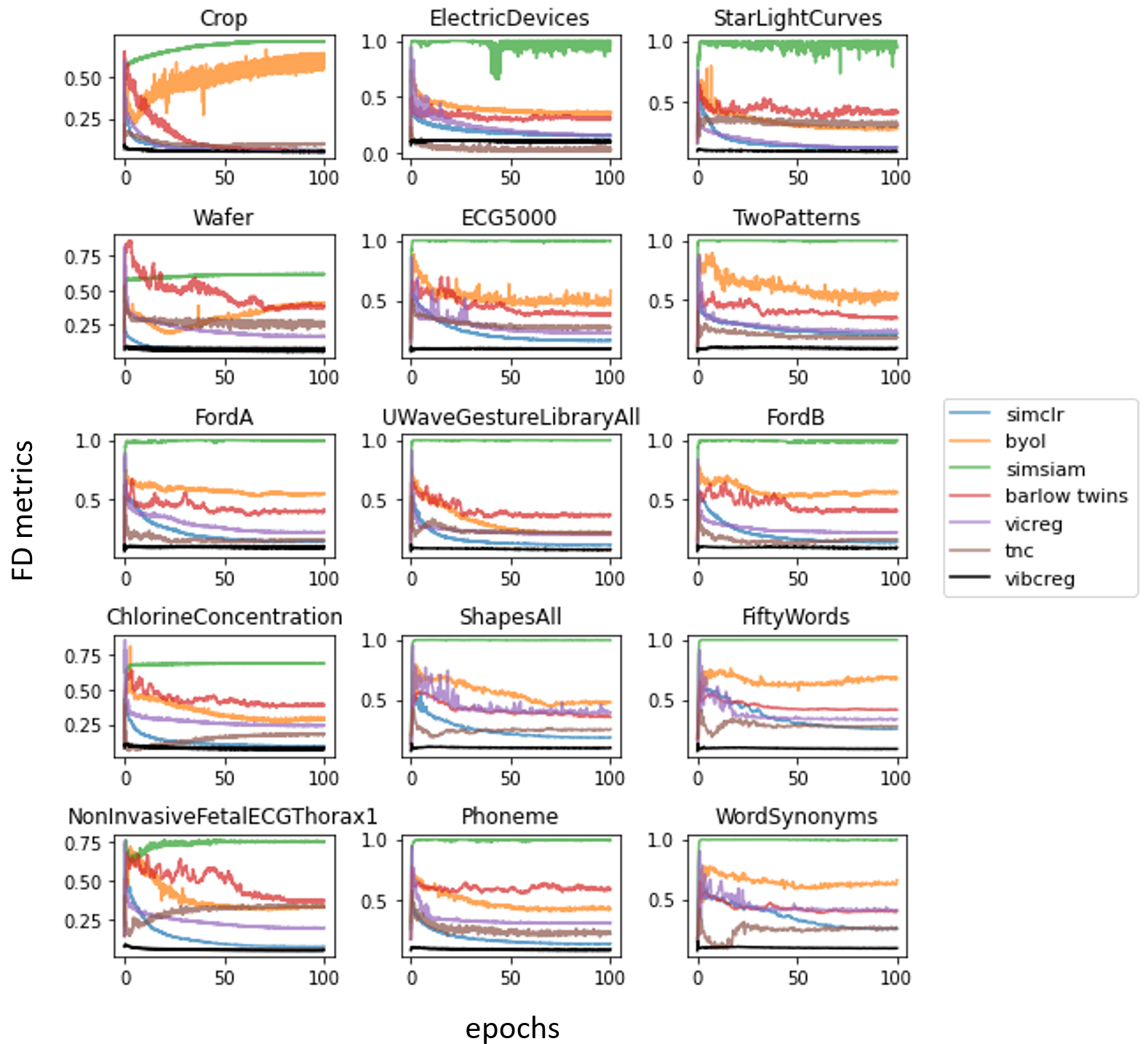}
    \caption{FD metric on the UCR datasets during the representation learning. It is noticeable that the metric of VIbCReg converges to near-zero in the beginning, ensuring the feature decorrleation throughout the entire training epochs.}
    \label{fig:UCR_FD_metrics}
\end{figure}

\begin{figure}[H]
    \centering
    \includegraphics[width=1.0\textwidth]{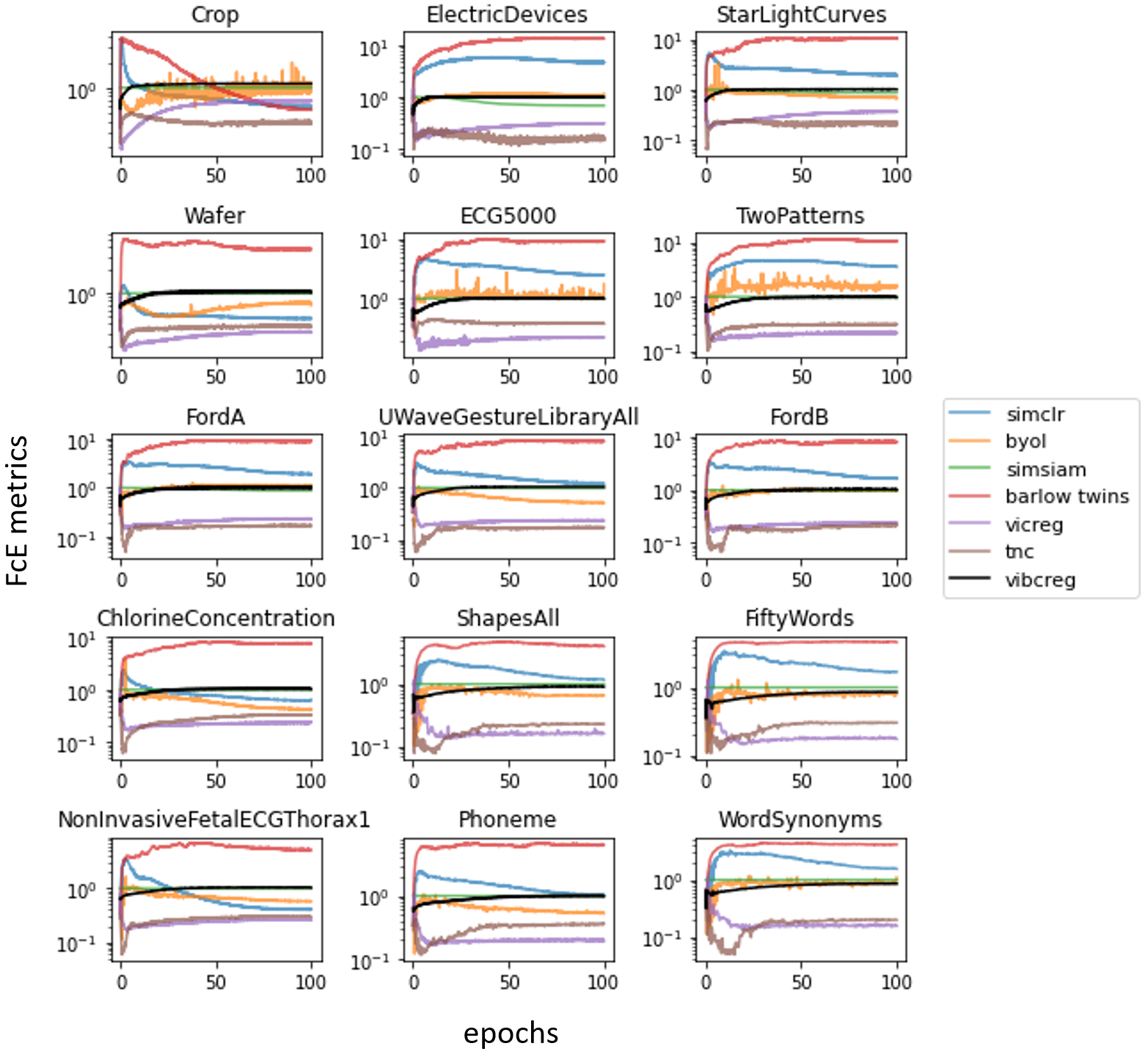}
    \caption{FcE metric on the UCR datasets during the representation learning. Note that y-axis is log-scaled. It can be observed that the metric of VIbCReg converges to 1 fairly quickly, ensuring the feature component expressiveness throughout the entire training epochs.}
    \label{fig:UCR_FcE_metrics}
\end{figure}

\paragraph{Faster Representation Learning by VIbCReg}
\label{appendix:faster_representation_learning_by_vibcreg}
The FD and FcE metrics that correspond to Fig. \ref{fig:UCR_knn_acc} (\textit{i.e., }5-kNN classification accuracy on the UCR datasets) are presented in Figs. \ref{fig:UCR_FD_metrics}-\ref{fig:UCR_FcE_metrics}, respectively. 

\end{appendices}

\bibliography{sn-bibliography}

\end{document}